\documentclass[lettersize,journal]{IEEEtran}
\usepackage{amsmath,amsfonts}
\usepackage{algorithmic}
\usepackage{array}

\usepackage{textcomp}
\usepackage{stfloats}
\usepackage{url}
\usepackage{verbatim}
\usepackage{graphicx}
\def\BibTeX{{\rm B\kern-.05em{\sc i\kern-.025em b}\kern-.08em
    T\kern-.1667em\lower.7ex\hbox{E}\kern-.125emX}}
\usepackage{balance}
\usepackage{soul}
\usepackage{color}
\usepackage{tabularx}
\usepackage{comment}

\usepackage[hang,small,bf]{caption}
\usepackage[subrefformat=parens]{subcaption}
\usepackage{booktabs}
\usepackage{makecell}
\usepackage{cite}

\begin{document}
\title{CROSS-Net: Region-Agnostic Taxi-Demand Prediction Using Feature Disentanglement \\
}

% \author{Ren Ozeki}
% % \affiliation{%
% %   \institution{Osaka University, Japan}
% %   \country{}
% % }
% % \email{r-ozeki@ist.osaka-u.ac.jp}

% \author{Haruki Yonekura}
% % \affiliation{%
% %   \institution{Osaka University, Japan}
% %   \country{}
% % }
% % \email{h-yonekura@ist.osaka-u.ac.jp}
% \author{Aidana Baimbetova}
% % \affiliation{%
% %   \institution{Osaka University, Japan}
%   \country{}
% }
% \email{a-baimbetova@ist.osaka-u.ac.jp}

% \author{Hamada Rizk}
% % \affiliation{%
% %   \institution{Tanta University, Egypt}
% %   % \institution{}
% %   \country{Osaka University,  Japan}
% % }
% % \email{hamada\_rizk@ist.osaka-u.ac.jp}

% \author{Hirozumi Yamaguchi}
% % \affiliation{%
% %   \institution{Osaka university, Japan}
%   \country{}
% }
% \email{h-yamagu@ist.osaka-u.ac.jp}

% \author{Ren Ozeki, Haruki Yonekura, Aidana Baimbetova, Hamada Rizk, Hirozumi Yamaguchi}

\author{
Ren~Ozeki,~\IEEEmembership{Student Member,~IEEE,}
Haruki~Yonekura,~\IEEEmembership{Student Member,~IEEE,}
Aidana~Baimbetova,~\IEEEmembership{}
Hamada~Rizk,~\IEEEmembership{Senior Member,~IEEE,}
and
Hirozumi~Yamaguchi,~\IEEEmembership{Member,~IEEE}
% \thanks{This work was partially funded by JST, CREST Grant JPMJCR21M5, and JSPS, KAKENHI Grant  22K12011, and NVIDIA award.}
% \thanks{R. O. is with is with the University of Osaka, Suita 565-0871, Japan (e-mail: r-ozeki@ist.osaka-u.ac.jp).}
% \thanks{H. Y. is with is with the University of Osaka, Suita 565-0871, Japan (e-mail: h-yonekura@ist.osaka-u.ac.jp).}
% \thanks{A. B. is with is with the University of Osaka, Suita 565-0871, Japan (e-mail: a-baimbetova@ist.osaka-u.ac.jp).}
% \thanks{H. R. is with 
% The University of Osaka, Suita 565-0871, Japan, and also with the Department of Computer and Control Engineering, Tanta University, Tanta 31733, Egypt (e-mail: hamada\_rizk@f-eng.tanta.edu.eg). }
% \thanks{H. Y. is with the University of Osaka, Suita 565-0871, Japan (e-mail: h-yamagu@ist.osaka-u.ac.jp).}}
}

% \markboth{IEEE SENSORS Journal,~Vol.~XXX, No.~XXX, month~2026}%
% {Ren Ozeki}

% \markboth{IEEE Transactions on Intelligent Transportation Systems,~Vol.~XX, No.~XX, Nov~2023}%

\def \sys {\textit{CROSS-Net}}

\IEEEpubid{}

\maketitle

\begin{abstract}
The growing demand for ride-hailing services has led to an increasing need for accurate taxi demand prediction. Existing systems are limited to specific regions, lacking generalizability to unseen areas. This paper presents a novel taxi demand prediction system, harnessing the strengths of multiview graph neural networks to capture spatial-temporal dependencies and patterns in urban environments. 
Additionally, the proposed system \sys\ employs a \textcolor{black}{spatially transferable} approach, enabling it to train a model that can be deployed to previously unseen regions.
To achieve this, the framework incorporates the power of a Variational Autoencoder to disentangle the input features into region-specific and region-agnostic components. The region-agnostic features facilitate cross-region taxi demand predictions, allowing the model to generalize well across different urban areas. Experimental results demonstrate the effectiveness of \sys \  in accurately forecasting taxi demand, even in previously unobserved regions, thus showcasing its potential for optimizing taxi services and improving transportation efficiency on a broader scale.
\end{abstract}


\section{Introduction}
% The increasing popularity of ride-hailing services has revolutionized urban transportation, providing convenient and efficient mobility options for individuals in bustling cities worldwide. However, the success of these services heavily relies on the ability to accurately predict taxi demand, allowing companies to effectively allocate resources, reduce customer waiting time, and enhance overall transportation efficiency. 

%\IEEEPARstart{T}{he}
The increasing popularity of ride-hailing services has revolutionized urban transportation. As of 2022, the global ride-hailing market was making revenue of approximately \$133 billion, with a growth rate of around 36\%, illustrating the widespread adoption of these services in cities across the globe \cite{Ride-hailing-Worldwide:online}. Such services promise to offer convenient and efficient mobility options for individuals in bustling cities worldwide. However, their optimal functioning and profitability hinge on the ability to accurately predict taxi demand. An accurate prediction ensures companies can effectively allocate resources, reduce customer waiting times, and save millions of dollars in operational costs. For instance, a 5\% improvement in demand prediction accuracy can translate to millions in savings for larger ride-hailing platforms~\cite{LIU2022100075}.

Traditional approaches to taxi demand prediction have predominantly relied on historical demand data and incorporating machine-learning techniques~\cite{Understanding_Taxi_Service_Strategies, xu2017real}.
However, these methods often suffer from limitations, primarily due to their inability to capture the complex spatial dependencies and patterns inherent in urban environments.
To capture spatial and temporal complex dependencies, the deep learning-based methods \cite{9439926, safikhani2020spatio} employ a CNN and an LSTM, resulting in improved accuracy compared to traditional methods.
However, since CNNs were designed for Euclidean space, such as images and grids, these methods have limitations in transportation networks with non-Euclidean topology and thus cannot essentially characterize the spatial correlation of traffic flow and road network.
In response to this challenge, Graph neural networks (GNNs) have gained considerable attention in recent years due to their ability to effectively model complex relationships in graph-structured data \cite{yao2018deep, Zhao_2020}.
By representing the city's road network as a graph—where nodes signify intersections or areas, and edges denote connectivity—GNNs can adeptly capture spatial relationships, even between geographically distant locations, as validated in~\cite{yao2018deep, Zhao_2020}.

% A significant limitation of existing taxi demand prediction systems is their lack of generalizability to unseen regions. Most models are trained and tailored to specific geographical areas, rendering them ineffective when applied to new or unobserved regions\cite{safikhani2020spatio, yan2020using}. 
% % This restricts the scalability and applicability of these systems, hindering their potential for broader deployment.
% Such lack of generalization limits the broader deployment of these systems across various urban landscapes, which are ever-evolving and unique in their dynamics. Crucially, the ability to generalize is not merely about versatility but also about scalability and efficiency. Without generalization, companies would face mounting costs and time expenditures each time they enter a new market or region, requiring new datasets and model retraining tailored to each specific locale. This limitation curbs rapid expansion, dampens responsiveness to emerging market opportunities, and hinders optimal service provision in diverse contexts.

Existing approaches often suffer from a lack of generalizability to unseen regions. These approaches, primarily tailored to specific geographical areas, underperform when deployed in new or unobserved regions \cite{safikhani2020spatio, yan2020using}. This issue exceeds mere versatility; it fundamentally impedes the scalability, efficiency, and broader deployment potential of these systems. 
Urban landscapes are characterized by their unique and evolving dynamics \cite{yao2018deep}. As each city has distinct transportation behaviors and patterns, an efficient prediction model must account for such variances. However, the current reliance on region-specific models creates a cascade of challenges: computational redundancies, resource-intensive recalibrations, and diminished deployment efficiency. Every new market entry would necessitate collecting region-specific datasets and retraining models, thereby increasing costs and time expenditures. Such constraints hinder rapid market expansion and timely response to emerging opportunities.
Furthermore, a notable technical challenge in current models is their tendency to overfit to compound region-dependent features, which implicitly represent an array of factors, including cultural, economic, mobility, structural, and spatio-temporal characteristics. Although such features are crucial for in-depth analysis within the training locale, their specificity excessively constrains the model, thereby diminishing its adaptability and jeopardizing its efficacy across varied geographies.

To tackle these challenges, we introduce \sys: a novel, region-agnostic taxi demand forecasting approach capable of generalization beyond spatial boundaries. It can be trained on one or more regions' data and can be applied effectively to any region, even those previously unobserved. We utilize a feature disentanglement strategy through a Variational Autoencoder (VAE) to disentangle the input features into region-specific and region-agnostic components. This separation allows the model to identify region-agnostic features that surpass geographical limits. Harnessing these region-agnostic features, our system can predict taxi demand across diverse regions, regardless of their region-specific attributes. Furthermore, \sys\ employs the strength of multi-view graph neural networks to discern spatial dependencies and patterns in urban settings. This approach not only ensures the scalability and practicality of the system but also maintains performance comparable to region-specific models.

% contribution  ~\cite{kipf2016variational}
\sys\ underwent a thorough evaluation using both a proprietary dataset and publicly available datasets to assess its demand prediction performance. The results validate the system's capability to achieve a remarkable taxi-demand prediction accuracy of at least 80.2\% in previously unseen regions across both datasets. This represents an improvement over existing state-of-the-art techniques, surpassing them by up to 28.6\%. These results demonstrate the feasibility of achieving predictive accuracy in real-world applications.

% {\color{blue}
% \textbf{Our contribution is fourfold:}
% 1) To our knowledge, this is the first work to address cross-region taxi demand prediction.
% \sys\ allows us to extract region-independent features for taxi demand, which guarantees the taxi demand prediction model of \sys\ is independent of regions.
% 2) Collection of real-world data from 20 different taxi providers in different cities and regions in Japan collected over a period of one year to evaluate \sys.
% 3) We conducted extensive experiments on real-world large-scale datasets. \sys\ achieved 80.2\% results and outperforms all other relevant state-of-the-art methods.
% We validate the individual contributions of each view Graph processing and highlight the importance of integrating all three views to achieve accurate taxi demand predictions.

% }

\textbf{Our contribution is fourfold:}
Firstly, to our knowledge, this is the first work to address cross-region taxi demand prediction.  \sys\ enables the extraction of \textcolor{black}{region-agnostic} features for taxi demand, ensuring the prediction model of \sys\ remains region-agnostic.
Secondly, incorporating a multiview graph processing module that discerns inherent relationships and dependencies among factors such as mobility trends, spatio-temporal demand patterns, and semantic features (e.g., land use). This allows for a holistic comprehension of both intra and inter-region dynamics.
Thirdly,  The collection of real-world data from 20 distinct taxi service providers across various cities and regions in Japan over a span of one year to evaluate \sys.
Lastly, the deployment of \sys\ to assess its efficacy in real-world scenarios and benchmark its performance using large-scale open datasets. \sys\ achieved at least 80.2\%  prediction accuracy in both datasets, surpassing the state-of-the-art approaches.

% This work also forms part of our broader research agenda on learning-based mobility analytics. In our previous studies, we investigated taxi demand prediction in decentralized environments through decentralized federated learning, and also explored privacy-related issues in transportation-oriented machine learning, including MIA-based privacy evaluation and Attribute Inference Attack-based vulnerability analysis\cite{ourMDMpaper, acm_src,ozeki2024privacy, gotoprivacy, matsumoto2023membership, 10.1145/3748636.3763226, Takeuchi2025_UrbComp, 10214923, yonekura2024restoring, 10723552}. These prior efforts provide a broader context for the present study and reflect our continuing interest in developing transportation prediction models that are accurate, reliable, and practically deployable.
% This work also aligns with our broader interest in domain generalization and location-agnostic learning in transportation and spatial computing. In our prior studies, we have explored cross-city autonomous driving and cross-region taxi-demand forecasting, together with building-agnostic and cross-building localization systems, all of which share the objective of enabling robust model transfer across heterogeneous environments. These efforts provide a broader context for the present study by highlighting our continuing interest in models that remain effective beyond a single city, region, or deployment setting\cite{ozeki2023onemodel, elkholy2023virtual, gaafar2025geodrive}.
This work also forms part of our broader research on learning-based mobility analytics. In our previous studies, we investigated taxi demand prediction in decentralized environments through decentralized and federated learning, explored privacy-related issues in transportation-oriented machine learning, including MIA-based privacy evaluation and attribute-inference-attack-based vulnerability analysis, and studied data-driven policy optimization for traffic management through simulation surrogation\cite{ourMDMpaper, acm_src,ozeki2024privacy, gotoprivacy, matsumoto2023membership, 10.1145/3748636.3763226, Takeuchi2025_UrbComp, 10214923, yonekura2024restoring, 10723552}.
In parallel, we have also studied domain generalization and location-agnostic learning in transportation and spatial computing, including cross-city autonomous driving, cross-region taxi-demand forecasting, and building-agnostic or cross-building localization systems\cite{ozeki2023onemodel, elkholy2023virtual, gaafar2025geodrive}. Together, these efforts reflect our continuing interest in developing models that remain accurate, robust, and practically deployable across heterogeneous environments.

This paper extends our earlier work on region-agnostic feature extraction for taxi demand prediction\cite{ozeki2023onemodel}.
In our previous work, we demonstrated the ability to accurately predict taxi demand in unseen regions using historical taxi demand data.
In this extended version, we have incorporated additional regional meta-information, such as land use and origin-destination (OD) statistics.
This work leverages multi-view graph processing, which allows our system to accurately predict taxi demand using three distinct perspectives: the spatio-temporal view, the mobility view, and the semantic view.
% This research validates the individual contributions of each view and highlights the importance of integrating all three views for achieving accurate taxi demand predictions.

This paper is organized as follows:
Section \ref{sec:preliminary} clarifies the problem statement and gives the definition. 
Section \ref{sec:sys_overview} presents our methodology, including graph representation, the architecture of our multi-view graph network, and the region-independent feature extraction.
Section \ref{sec:evaluation} discusses the experimental setup, and analyzes the results.
Section \ref{sec:related_work} reviews related work in taxi demand prediction and graph-based models.
Finally, Section \ref{sec:conclusion} concludes the paper.

\color{black}
\section{Preliminary and Problem Definition} \label{sec:preliminary}
In this section, we introduce some definitions and then formalize the taxi demand prediction problem.
\subsection{Definitions}
\label{sec:definitions}
\begin{itemize}
    \item \textbf{Region:}
     A spatial region, denoted as $r \in \textbf{R}$, pertains to a well-defined area within a geographical context. It comprehensively encompasses diverse administrative divisions, including prefectures, cities, and specific regions within a city. To illustrate, we can examine the city of Tokyo as an exemplar of a spatial region. This particular spatial region can be denoted as $r_{\text{Tokyo}}$, representing the entirety of Tokyo's area, which encompasses its administrative divisions, neighborhoods, landmarks, and the surrounding regions.

    \item \textbf{Cell:}
    % spot, location, area, etc.
    A cell, within the context of the taxi demand prediction task, is defined as the smallest unit on which the region is partitioned. The objective is to train a taxi demand system to make predictions at this granular level.

    \item \textbf{Time slot:}
    % The continuous time is partitioned into sequential and equal time intervals. Each slot is denoted as \textcolor{blue}{$T_k$, $1 \leq k \leq K$}, where $K$ represents the number of time intervals during a continuous period of time. For instance, if a time interval of 30 minutes is set, one week which has 7 days can be divided into $48 \times 7$ slots, and $K$ is set to $336$.
    The continuous time is partitioned into sequential and equal time intervals. The suffix $t$ of \textit{time slot} indicates the duration from a point in time to the next point in time. We set a time interval of 30 minutes to provide a balance between granularity and computational efficiency, allowing for accurate predictions while maintaining efficient processing for taxi demand systems.

    \item \textbf{Taxi Demand:}
    In the region $r$, passengers locally request or pick up a taxi and travel to their destination. The number of taxi demands in each \textit{cell} is defined as the sum of the number of pick-ups in each \textit{time slot}.

    \color{black}
    % \item \textbf{Spatial transferability:}
    % In transportation forecasting, a key practical requirement is \textit{spatial transferability}, namely the ability to apply a model developed in one region to another region without rebuilding the model from scratch.

    \item \textbf{Region-agnostic:}
    In this paper, we use the term \textit{region-agnostic} to denote \textit{cross-city generalization}. A region-agnostic model is expected to maintain high prediction performance when deployed to a previously unseen region under the leave-one-city-out setting.

    \item \textbf{Region-specific:}
    We use \textit{region-specific} to describe factors that are tied to a particular urban context, such as region-dependent demand scales and mobility structures.
    
    \color{black}
    
\end{itemize}

\begin{figure*}[tb]
  \begin{tabular}{c}
    \begin{minipage}{0.23\linewidth}
      \begin{center}
      \includegraphics[width=\linewidth]{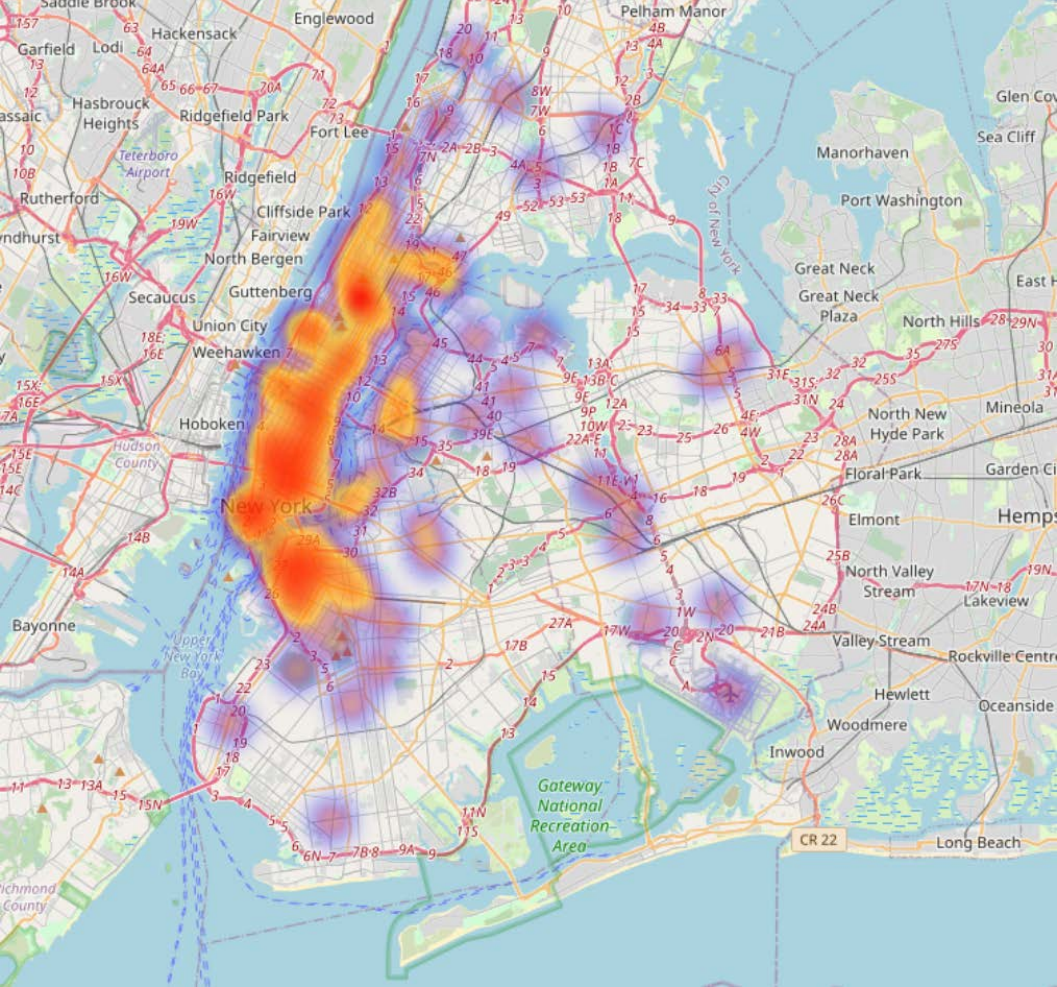}
      \end{center}
      \subcaption{From 0 a.m. to 1 a.m..}
      \label{fig:NY_00}
    \end{minipage}

    \begin{minipage}{0.23\linewidth}
      \begin{center}
      \includegraphics[width=\linewidth]{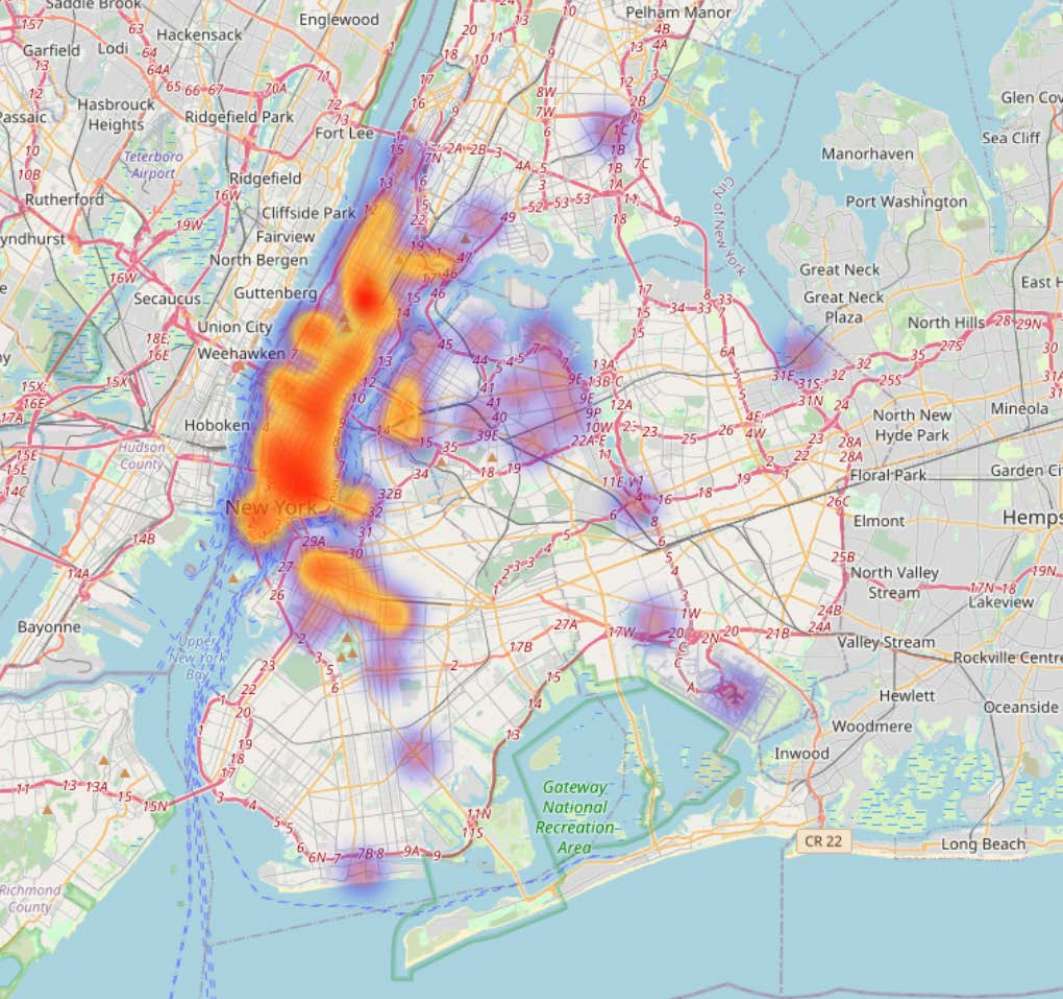}
      \end{center}
      \subcaption{From 3 a.m. to 4 a.m..}
      \label{fig:NY_03}
    \end{minipage}

    \begin{minipage}{0.23\linewidth}
      \begin{center}
      \includegraphics[width=\linewidth]{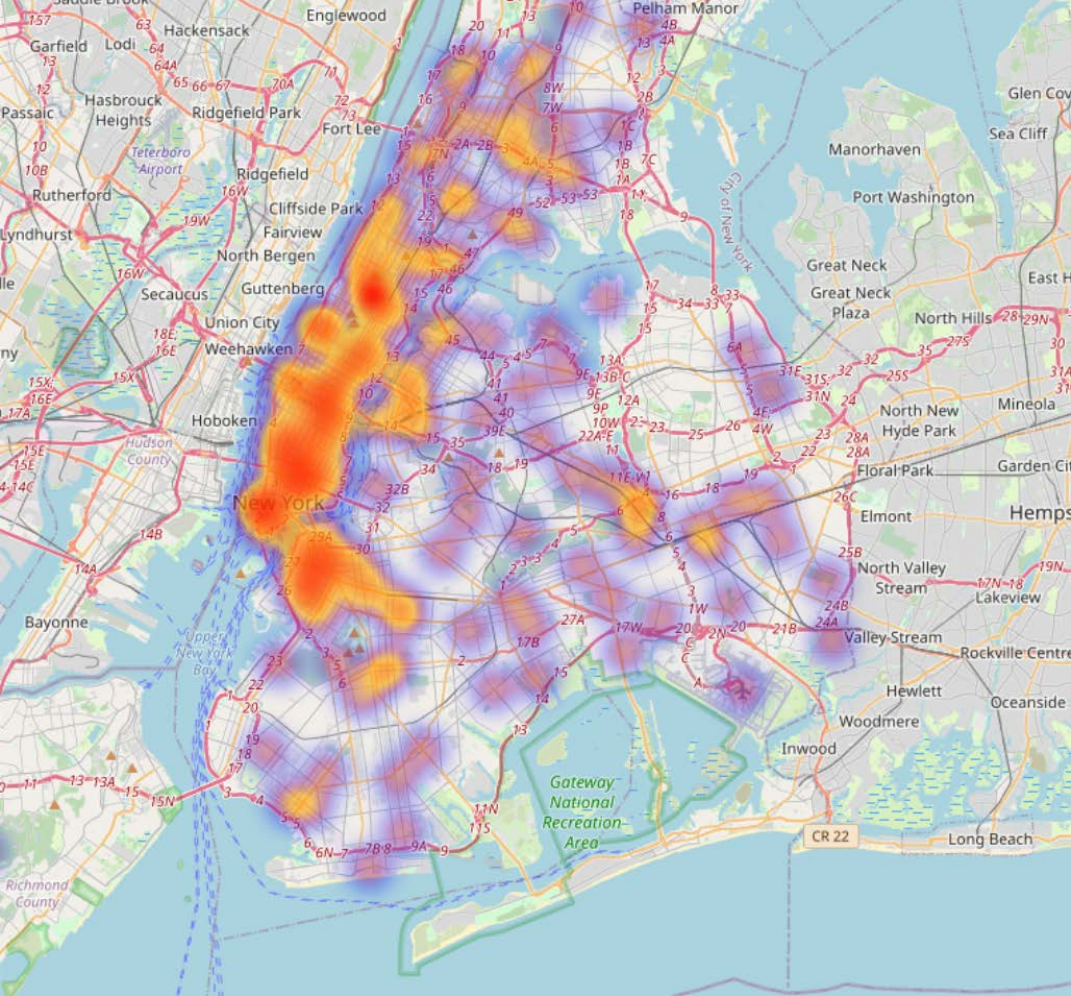}
      \end{center}
      \subcaption{From 6 a.m. to 7 a.m.}
      \label{fig:NY_06}
    \end{minipage}

    \begin{minipage}{0.23\linewidth}
      \begin{center}
      \includegraphics[width=\linewidth]{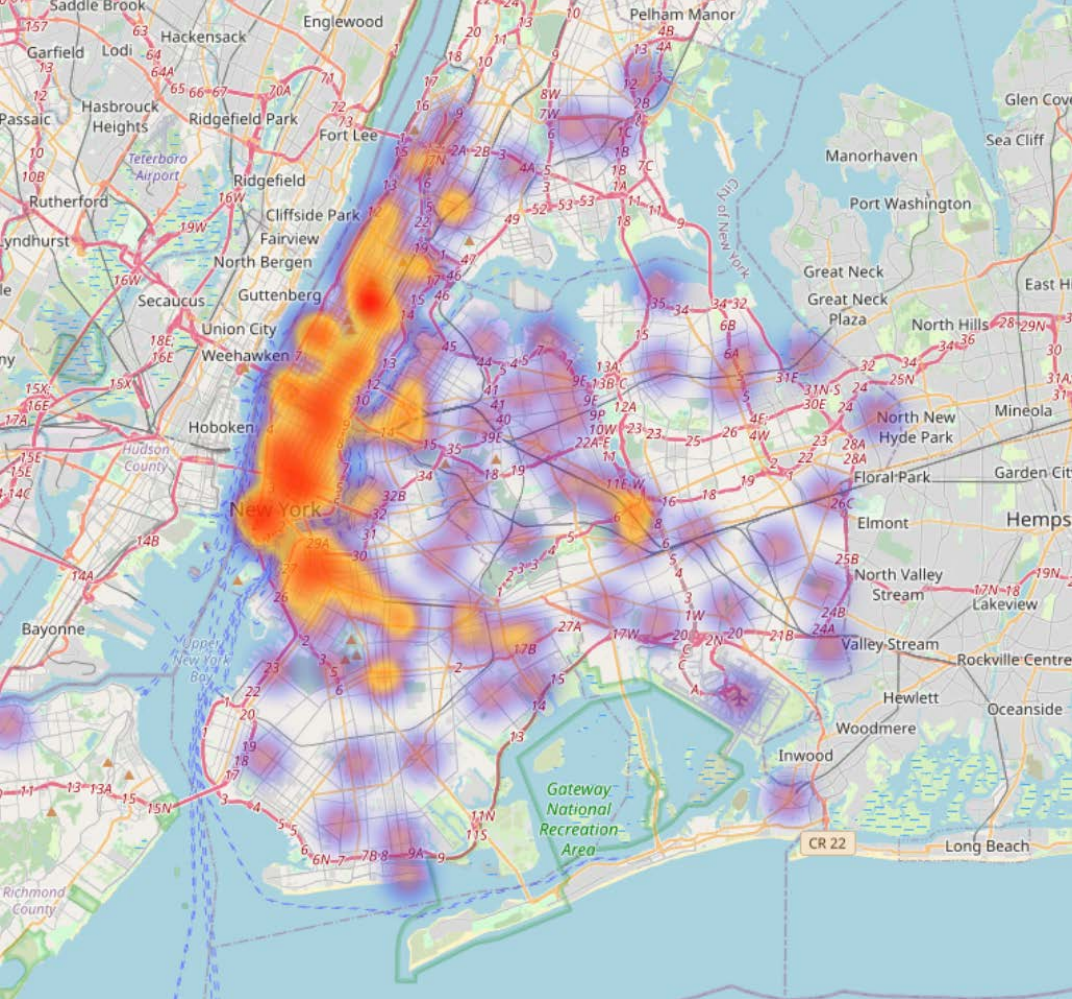}
      \end{center}
      \subcaption{From 9 a.m. to 10 a.m.}
      \label{fig:NY_09}
    \end{minipage} \

  \end{tabular}
  \caption{Demand distribution heat-map in New York City.}
  \label{fig:snapshotsNY}
\end{figure*}

\begin{figure*}[tb]
  \begin{tabular}{c}
    \begin{minipage}{0.23\linewidth}
      \begin{center}
      \includegraphics[width=\linewidth]{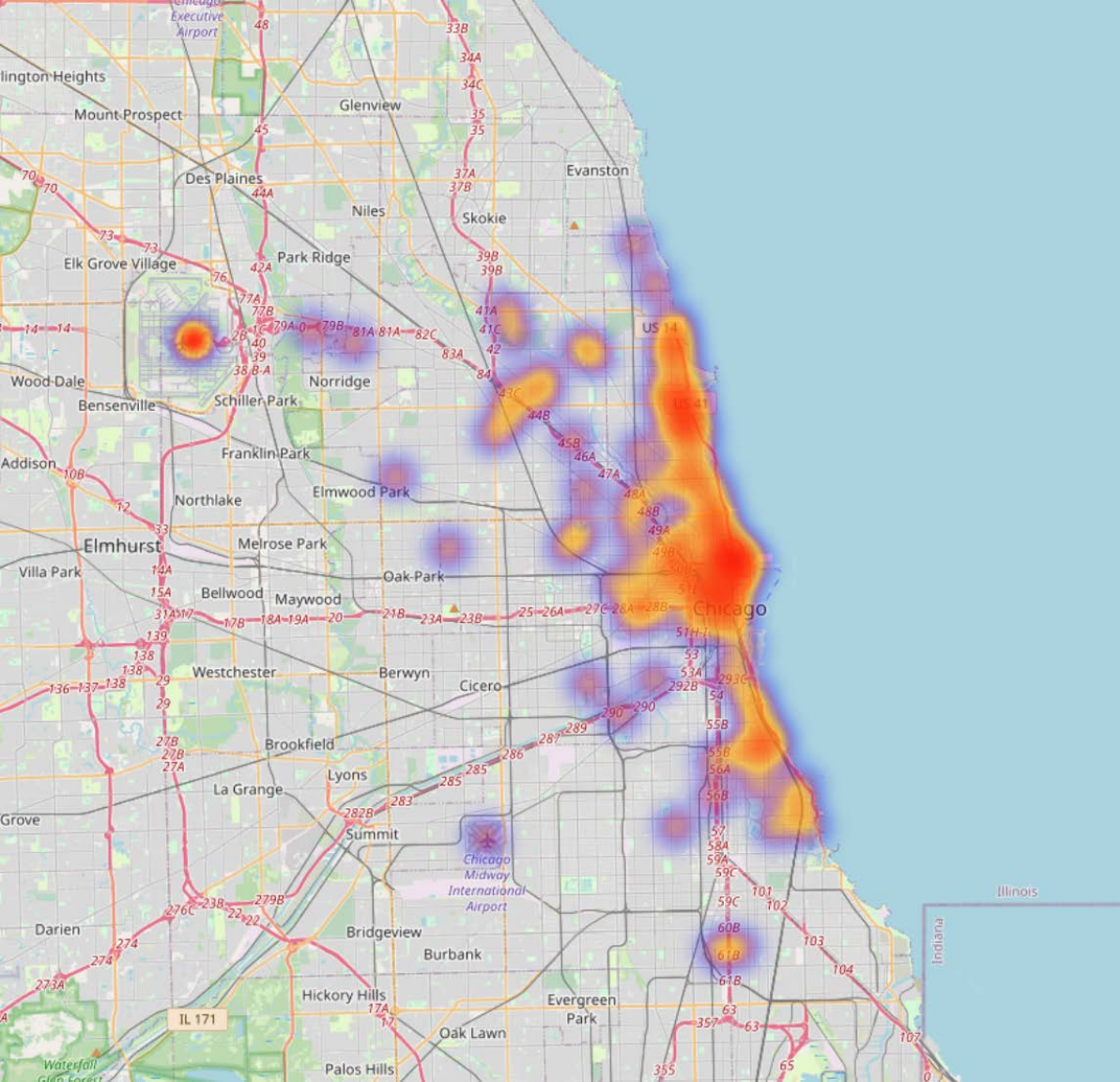}
      \end{center}
      \subcaption{From 0 a.m. to 1 a.m..}
      \label{fig:chicago_00}
    \end{minipage}

    \begin{minipage}{0.23\linewidth}
      \begin{center}
      \includegraphics[width=\linewidth]{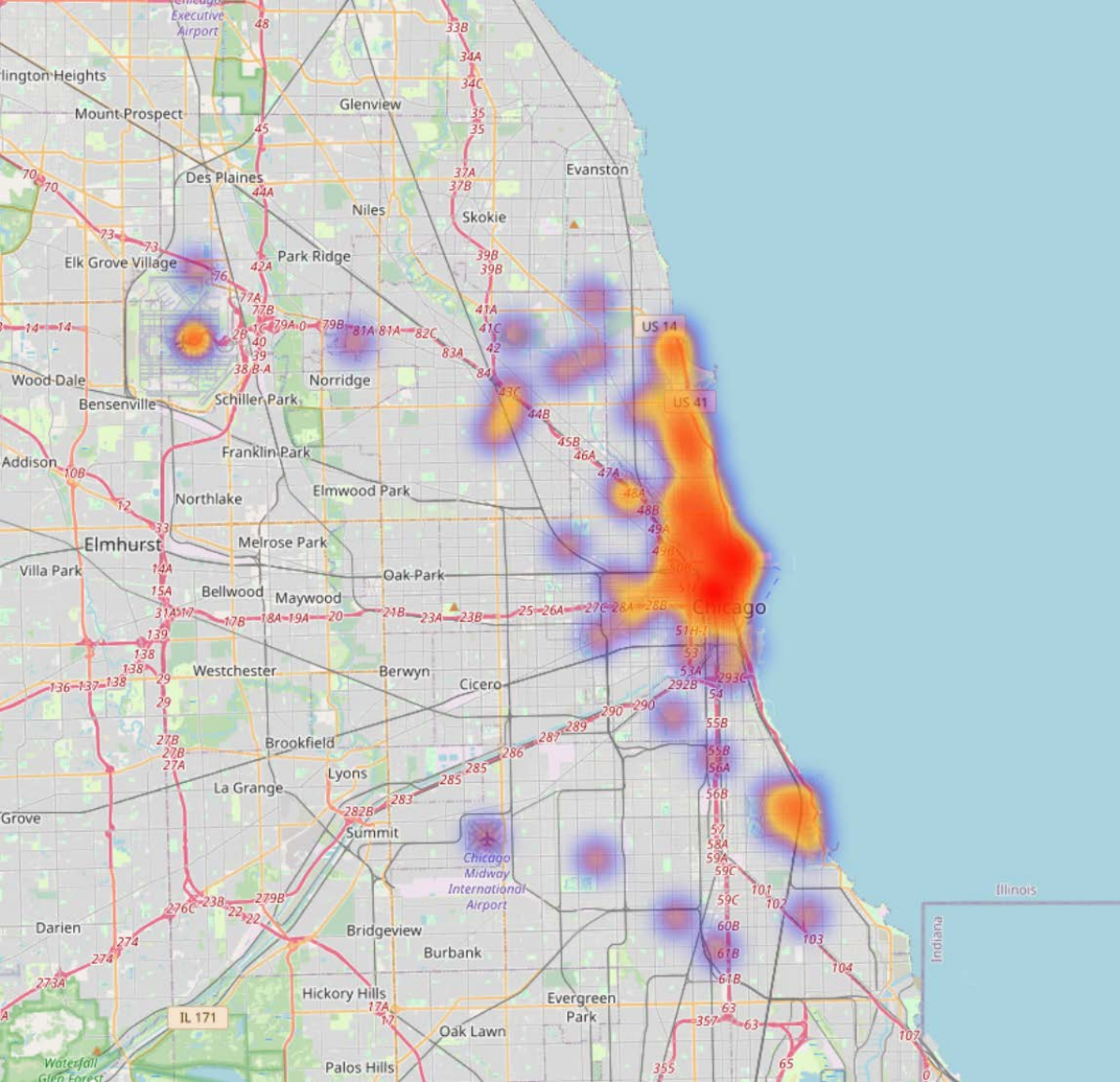}
      \end{center}
      \subcaption{From 3 a.m. to 4 a.m..}
      \label{fig:chicago_03}
    \end{minipage}

    \begin{minipage}{0.23\linewidth}
      \begin{center}
      \includegraphics[width=\linewidth]{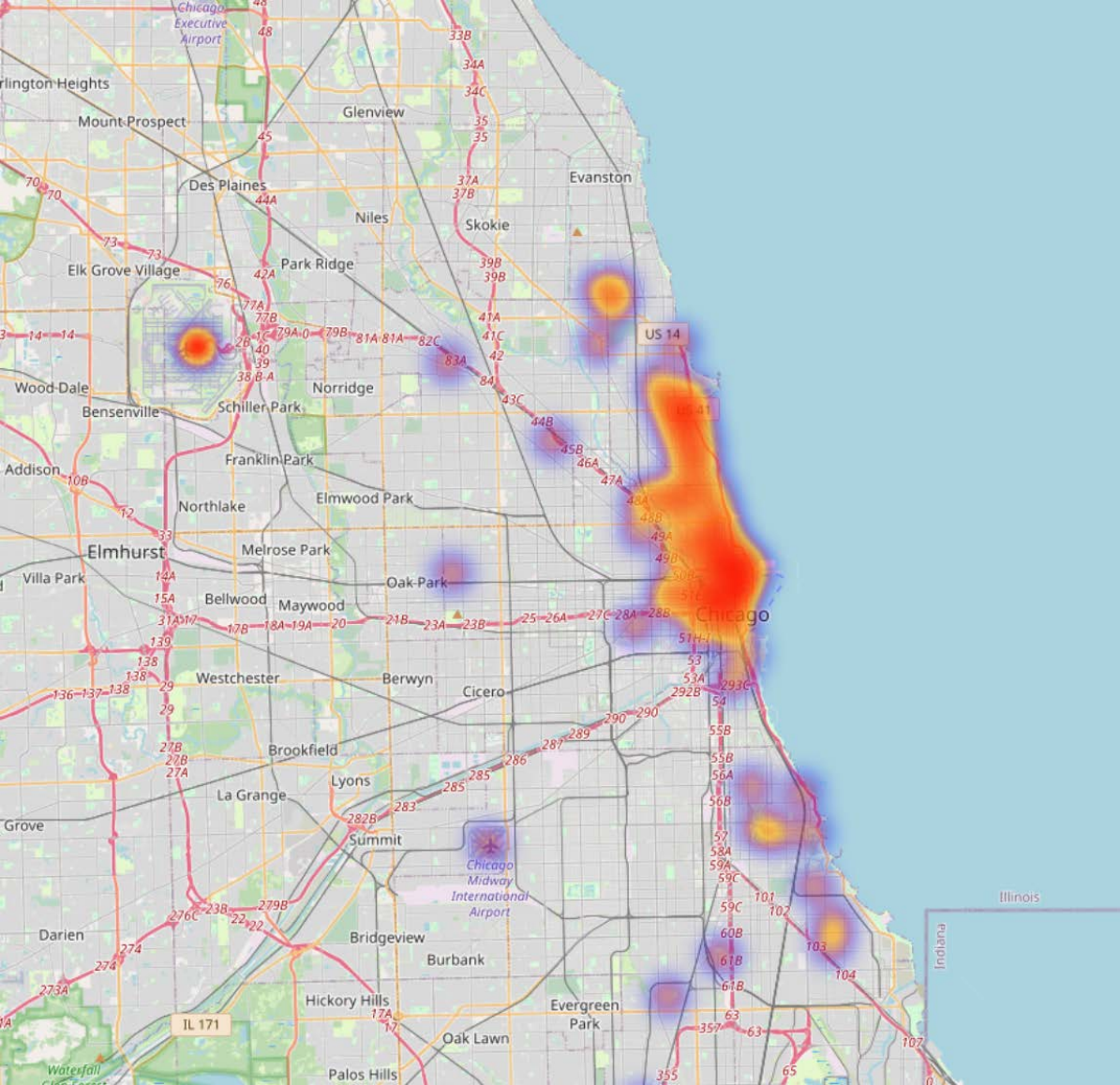}
      \end{center}
      \subcaption{From 6 a.m. to 7 a.m.}
      \label{fig:chicago_06}
    \end{minipage}

    \begin{minipage}{0.23\linewidth}
      \begin{center}
      \includegraphics[width=\linewidth]{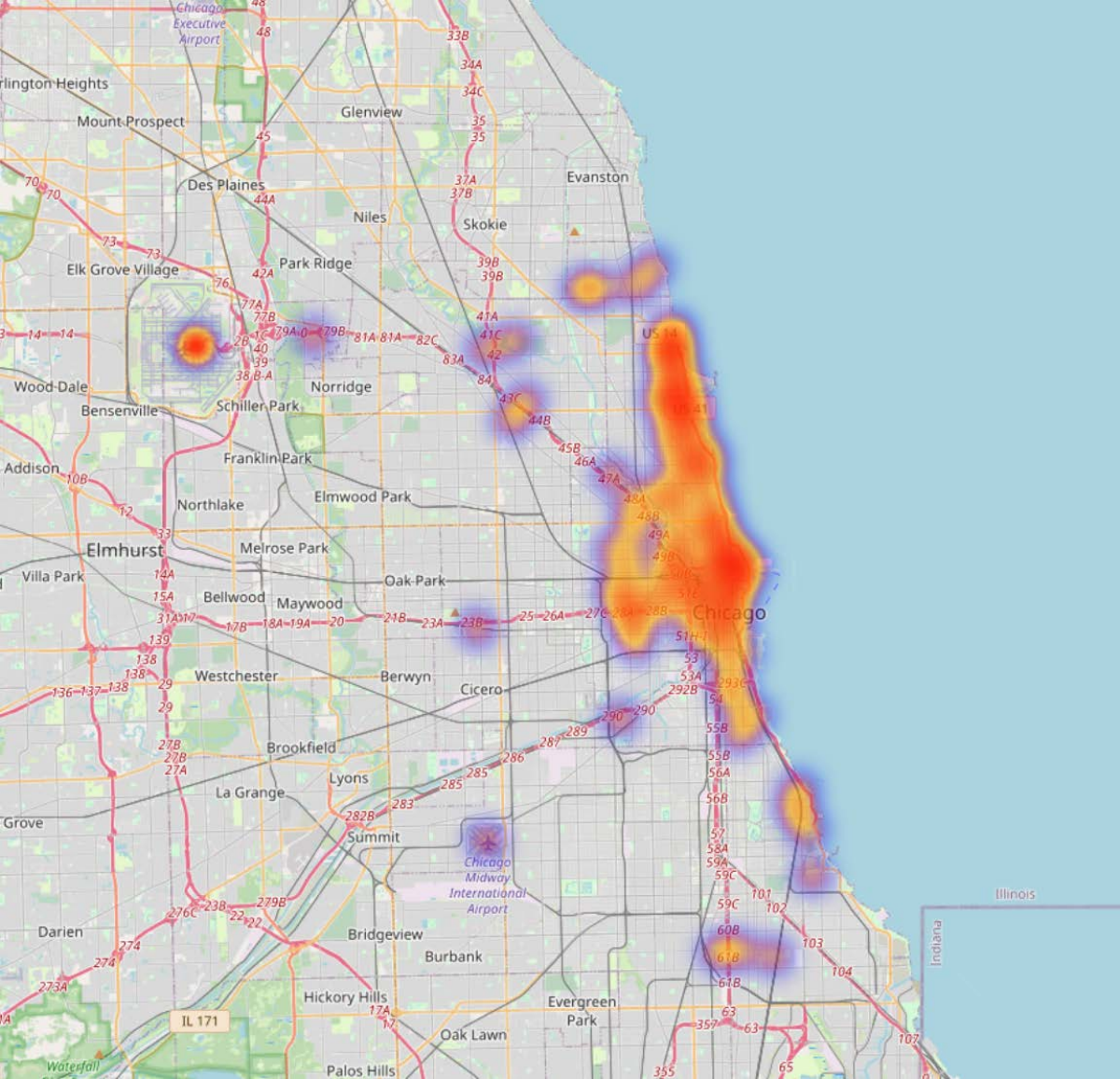}
      \end{center}
      \subcaption{From 9 a.m. to 10 a.m.}
      \label{fig:chicago_09}
    \end{minipage} \

  \end{tabular}
  \caption{Demand distribution heat-map in Chicago.}
  \label{fig:snapshotsChicago}
\end{figure*}

\begin{figure}[tb]
    \centering
    \includegraphics[width=7.5cm]{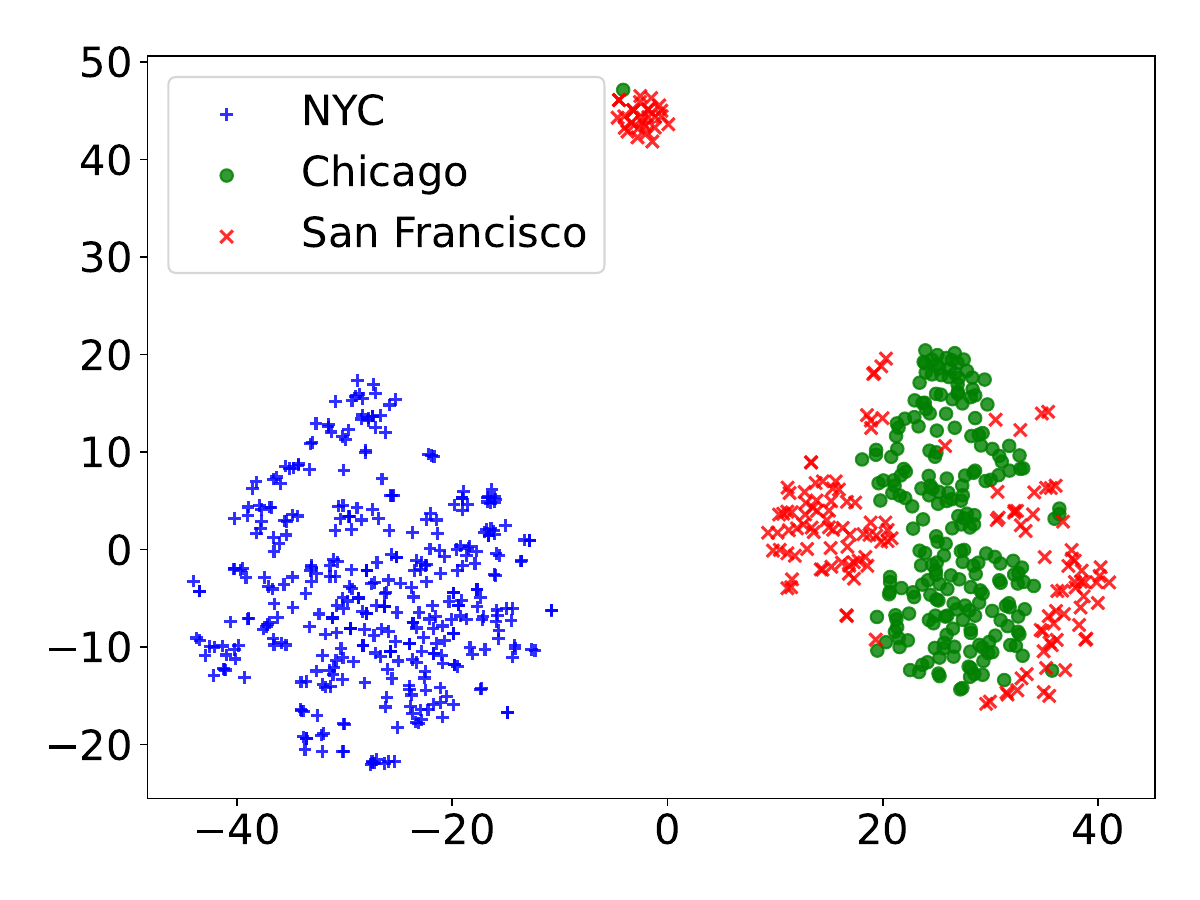}
    \caption{t-SNE of original features of 3 regions, NYC, Chicago, and San Francisco.}
    \label{fig:tsne_sample}
\end{figure}

\begin{figure}[tb]
    \centering
    \includegraphics[width=7.5cm]{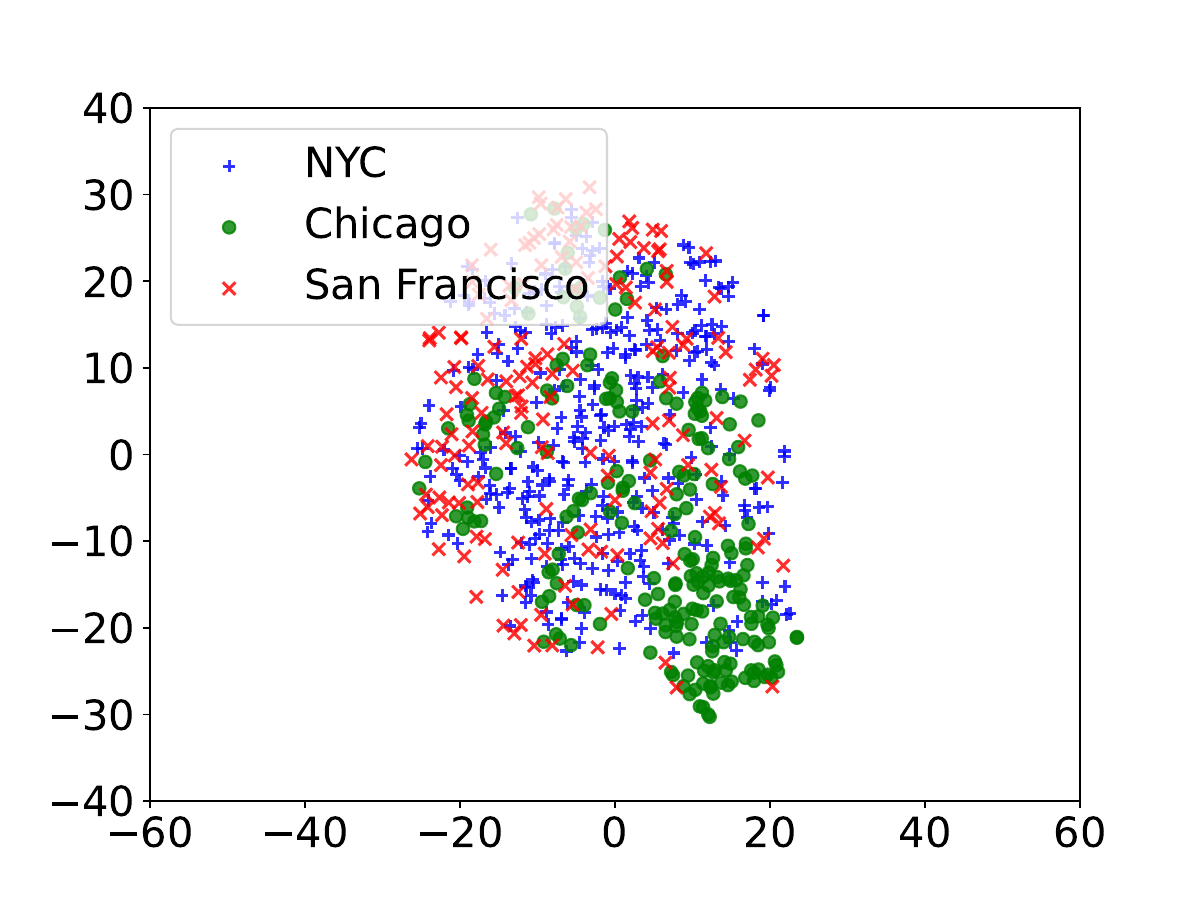}
    \caption{t-SNE of region-agnostic representation of 3 regions, NYC, Chicago, and San Francisco.}
    \label{fig:tsne_sample_region_neutral}
\end{figure}

\color{black}
\subsection{Problem Formulation}
\label{sec:Problem_Fomalization}

\begin{table}[t]
    \centering
    \caption{Notations used in the paper.}
    \label{tab:notation_table}
    \resizebox{\linewidth}{!}{%
    \begin{tabular}{ll}
    \hline
    Notation & Description \\
    \hline
        $\textbf{R}$ & Region index space \\
        $r$ & Region $r \in \textbf{R}$ \\
        $\textbf{X}$ & Feature value for taxi demand\\
        $\textbf{X}^{r}$ & Feature value of Region $r$\\
        $\textbf{Y}$ & Taxi demand space\\
        $y$ & The amount of demand $\in \textbf{Y}$\\
        $\mathcal{F}$ & Function assigning each $\textbf{X}$ to one element $y$ \\
        $t$ & Index of \textit{time slot} \\
        $i$ & Index for cell\\
        $e^{i}_t$ & External features for a cell $i$ at time $t$ \\
        $G$ & Graph representing some information in a region \\
        $D_t$ & Demand matrix at time $t$ \\
        $h$ & Output channel of the graph processing of each view \\
        $x^{r}$ & Latent input feature for Region \textit{r} \\
        $z_{r}$ & Region specific feature\\
        $p(\cdot)$ & Decoder in the region agnostic module \\
        $q(\cdot)$ & Encoder in the region agnostic module \\
        $\Phi$ & Parameters of the encoder of $q(\cdot)$ \\
        $SR_y$ & Domain prediction module \\
        $SC_r$ & Region classifier \\
        $\ell (\cdot)$ & Cross entropy function\\
        $z$ & Region agnostic feature\\
        $L_{\cdot}$ & Loss function installed in the region agnostic module \\
        $\alpha, \beta$ & Hyperparameters in loss function of the region agnostic module \\
    \hline
    \end{tabular}
    }
\end{table}

Taxi demand prediction is a problem that aims to predict the demand at time slot $t+1$, given the data until time slot $t$.
In addition to historical demand data, we can also use other statistical and meta-features of the region such as mobility statistics, Points of Interest
(POI), and meteorological data.
We define those external features for a cell $i$ and time step $t$ as a vector $e^{i}_{t} \in \mathbb{R}^{l}$, where $l$ is the number of features. 
Therefore, our target application is formulated as eq.\ref{eq:taxi_dmeand_prediction}.

\begin{equation}
\label{eq:taxi_dmeand_prediction}
    y_{t+1}^{i} = \mathcal{F}(X_{i})= \mathcal{F}(y^{i}_{t-h,\cdots,t}, e^{i}_{t-h,\cdots,t})
\end{equation}

where $y^{i}_{t-h,\cdots,t}$ is historical demand, $e^{i}_{t-h,\cdots,t}$ is an external feature for a cell $i$ from the time slot $t-h$ to $t$, and $h$ is a fixed number of preceding time slots (historical).
Our notation is summarized in table~\ref{tab:notation_table}.

\begin{figure*}[tb]
% %\vspace{-0.5cm}
    \centering
    \includegraphics[width=17cm]{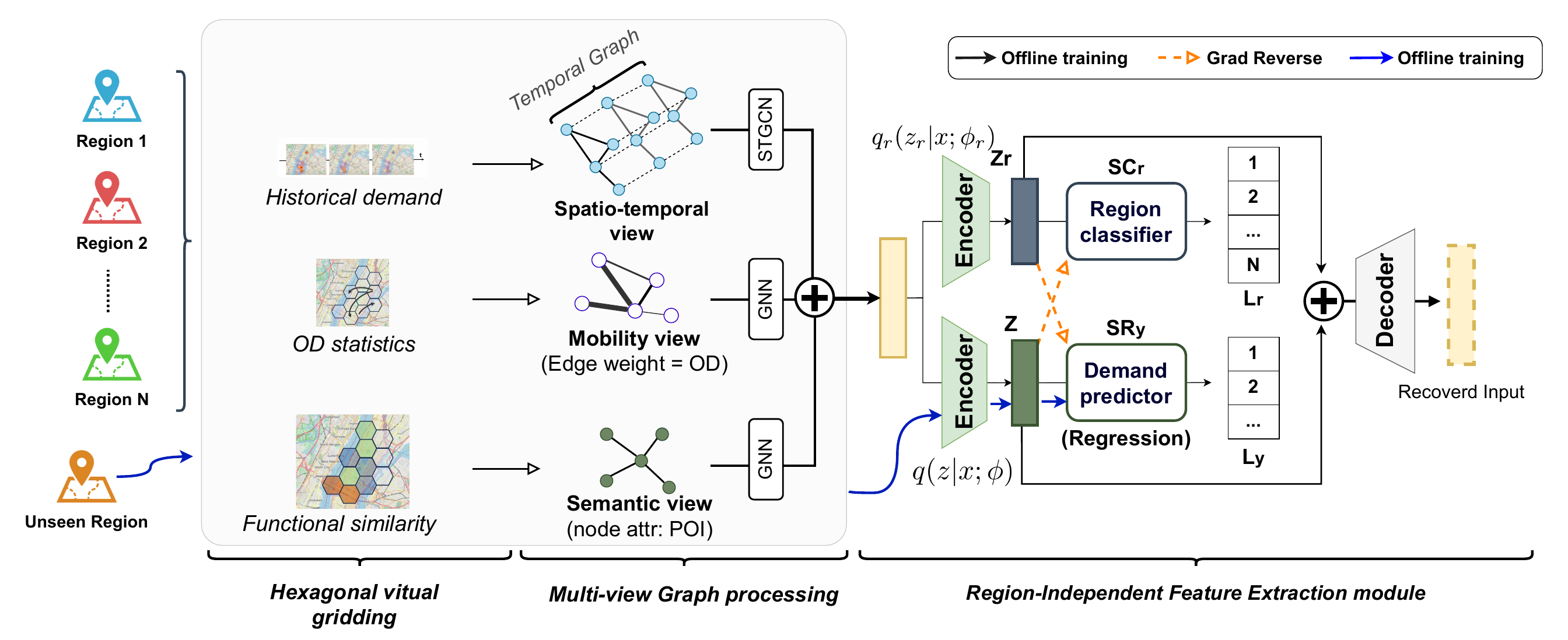}
    \caption{Proposed framework overview. The black line represents the training phase, and the blue line represents the run-time procedure. Our framework aims to separate region-specific feature $z_r$ and region agnostics feature $z$ from multi-view graph input features using two encoders.}
    \label{fig:system_architecture}
\end{figure*}

In this paper, 
% we define regions $r \in \textbf{R}_{train} = \{1, \cdots N\}$. %where $\textbf{R}$ denotes the set of regions.
our goal is to train a deep-learning model that extracts region-independent representation to predict taxi demand in unseen regions.
Specifically,  given labeled data from N source regions $\{(X^{r}, Y^{r})\}^{N}_{r=1}$ where $X^{r}$ is set of feature in region $r$ and $Y^{r}$ is the demand for region $r$.
% where $X^{r} = {}_{r}y^{i}_{t-h,\cdots,t}, {}_{r}e^{i}_{t-h,\cdots,t}$
To make sure that the representation is independent of region-specific knowledge, we evaluate the prediction model by the test data $X^{\tilde{r}}$ from the unseen region, where $\tilde{r}\notin \textbf{R}_{source}$, where $\textbf{R}_{source}$ is the set of regions used for training the model.

The primary challenge in predicting taxi demand for unseen regions arises from the substantial variations in demand patterns across different regions, as depicted in Fig.~\ref{fig:snapshotsNY} and Fig.~\ref{fig:snapshotsChicago}. Several factors influence these disparities, such as spatial distribution, temporal trends, and geographic characteristics. Fig.~\ref{fig:tsne_sample} offers a t-SNE representation, emphasizing the complexity of this issue. The t-SNE embedding of the original features in Fig.~\ref{fig:tsne_sample} clearly demonstrates the separation of data points based on their geographical regions.
This visualization underscores that the taxi demand prediction model $\mathcal{F}^{r_{Chicago}}$, trained using data from region $r_{Chicago}$ (green dots), struggles to accurately predict the demand $Y^{r_{NYC}}$ for region $r_{NYC}$ when provided with inputs $X^{r_{NYC}}$ (blue dots). This performance limitation arises from the challenge posed by out-of-distribution data \cite{PAWAR2022379, shen2021towards}.
Conversely, when the model is trained using region-agnostic features, as depicted in Figure~\ref{fig:tsne_sample_region_neutral}, it demonstrates the ability to perform well in previously unseen regions. This observation highlights the importance of extracting region-independent feature representations to achieve robust and general taxi demand prediction.

Our primary objective is to achieve region-agnostic taxi demand prediction by leveraging a deep learning framework that can effectively transform region-specific features into region-independent representations, as exemplified in Figure~\ref{fig:tsne_sample_region_neutral}. By training this framework using data from the source regions, we aim to create a model that can accurately predict demand in any target region. This unified model eliminates the need for region-specific models and enables the utilization of a single model for predicting demand across various regions.

\section{The Proposed Framework} \label{sec:sys_overview}
\color{black}

\subsection{Framework Overview}
\sys, as illustrated in Figure~\ref{fig:system_architecture}, operates in two stages: an offline training stage and an online inference stage.
During the offline training stage, the framework leverages a large dataset consisting of labeled samples collected from diverse regions. These samples are characterized by cell information, timeslots, and semantics of location. Then the \textbf{Hexagonal Virtual Gridding} module partitions the map into hexagonal cells and calculates taxi demand for each cell and timeslot, transforming the labeled samples into an interpretable format suitable for machine learning models.
These spatio-temporal data are then represented as a multi-view graph structure by the \textbf{Graph Processing} module. This module captures the inherent relationships and dependencies among the features, enabling a comprehensive understanding of the region's dynamics.
Following this, the multi-view graph is inputted into the \textbf{Region-Independent Feature Extraction} module, which facilitates the extraction of region-agnostic and region-specific features separately. This step ensures a clear distinction between factors that are influenced by the region itself and those that are independent of the region.
Lastly, the framework trains the \textbf{Taxi Demand Prediction Model} using the region-agnostic features, enabling the model to generalize and make accurate predictions across different regions.

In the online stage, the framework enables taxi service providers to query the system for forecasting demand patterns in any region at specific time intervals, even those previously unobserved. This involves processing associated views through the graph processing module to construct a multi-view graph representing the region. The resulting graph is then passed to the pre-trained encoder model to extract region-independent features. These features are subsequently utilized by the unified taxi demand prediction model to forecast the demand within the target region.

\begin{figure*}[tb]
    \centering
    \includegraphics[width=17cm]{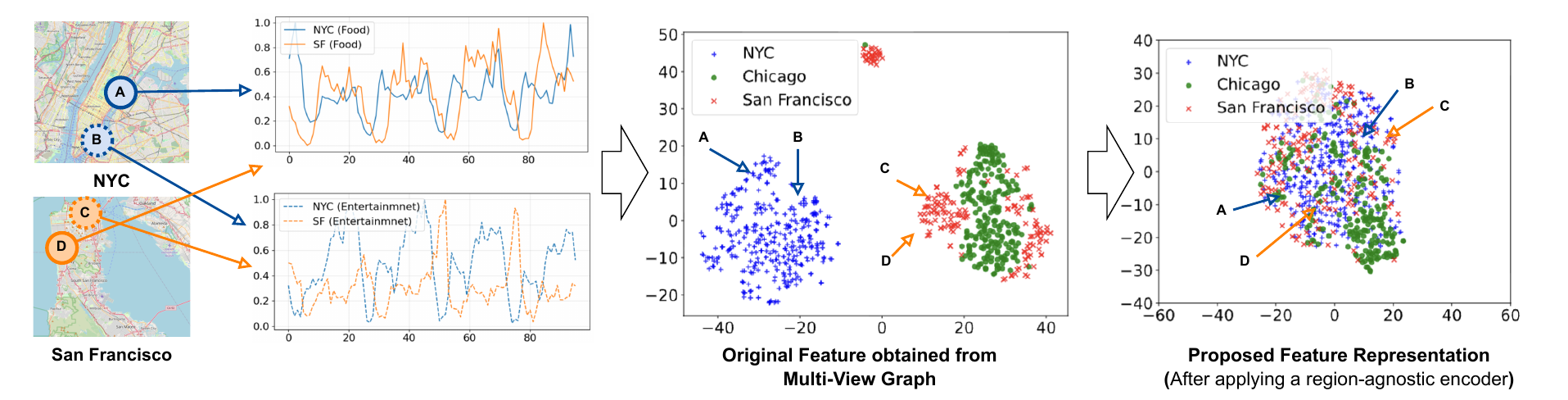}
    \caption{\textcolor{black}{Entangled latent representations obtained from multi-view graph are transformed into region-agnostic features through the proposed disentanglement approach.}}
    \label{fig:feature_extraction_with_geo_visualization}
\end{figure*}

\color{black}
\subsection{Hexagonal Virtual Gridding}\label{sec:hex_grid}
This module is a pivotal component of the data generation system aimed at accurately computing the taxi demand for inputs of machine learning models. This module plays a vital role in transforming raw trajectory data into a manageable and interpretable format suitable for utilization by machine learning models.
The primary objective of this module is to partition the map into evenly spaced hexagonal cells, wherein each cell represents a distinct area on the map. Subsequently, the module calculates the number of demand events that transpired within each hexagonal cell during a specific time-slot in a day. Notably, the module does not differentiate between pick-up and drop-off events, solely focusing on the aggregate count of demand events within each cell.
The gridding process entails superimposing a virtual hexagonal grid onto the map. This approach enables the system to provide a comprehensive overview of taxi demand in different areas of the city, facilitating predictions regarding the number of demand events in distinct areas. Moreover, it allows for the facile visualization of demand patterns and the identification of regions characterized by high or low demand.

We adopt a hexagonal grid as opposed to a square grid due to its efficiency and effectiveness in representing geographic regions compared to squares. Specifically, hexagons provide balanced neighboring as each hexagon shares a common edge with six neighboring hexagons. This property ensures a more equitable distribution of neighboring cells, edge-reducing effects and accurately representing spatial relationships. Additionally, hexagons allow for more compact packing, covering a given area with fewer cells, resulting in a more accurate depiction of the geographic space and reduced redundancy within the grid. 
Moreover, hexagonal cells have equidistant centers, ensuring consistent and regular spacing throughout the grid. This characteristic facilitates precise distance calculations and enables robust spatial analysis. 
Furthermore, hexagons offer directional flexibility, allowing movement in six possible directions. 
This flexibility enhances the system's ability to capture and analyze spatial patterns, making it particularly advantageous in modeling transportation systems and understanding travel patterns.
\color{black}
\subsection{Graph Processing}
\label{sec:graph_proc}

% \hl{Yonekura: make node feature, edge and edge weight clear}
This module is responsible for representing the input spatio-temporal features in multi-view graph structures. This is due to the ability of graph representation to capture and represent complex relationships and dependencies among various components of the data, which enhances the effectiveness of predicting taxi demand, even in previously unobserved regions.
To achieve this, we employ three distinct types of graphs as input: the historical demand graph, mobility graph, and region meta-information graph. Each of these graphs serves a specific purpose in capturing relevant information and facilitating the prediction process. The historical demand graph captures past demand trends and spatial distributions, providing valuable insights for predicting future demand. The mobility graph represents the mobility trends within the region, which are important factors for demand prediction. Lastly, the region meta-information graph incorporates semantic information, allowing cells with similar functionality to be associated with similar demand patterns.

Generally, each graph consists of nodes and edges. Each node encodes spatio-temporal information, such as the relative latitude/longitude of the containing cell and the day of the week, which are represented as one-hot vectors, and the time slot.
%Additionally, the time interval is set by a resolution parameter $\alpha$ (e.g., $\alpha$ = 1-hour results in 24 slots).
% time interval : <- sequence of historical graph, consisting of multiple time sk
% time slot : <- for demand, 
The edges connect nodes that are spatially related, such as adjacent cells, and the weight of each edge corresponds to the connectivity of the nodes in terms of vehicle mobility. The weight is determined by the volume of traffic during a specific time period, indicating the strength of connectivity between nodes on the map.

\subsubsection{Spatio-temporal view}
\label{sec:st-view}
Since taxi demand strongly depends on spatio-temporal features, past demand trends, and spatial distributions are useful to predict future taxi demand\cite{chen2020multitask, yao2018deep}.
For every time slot, we consider one cell as a node and a region as a graph that represents the taxi demand map.
The graph of the demand map has one node feature, demand value, and edges that connect adjacent cells.
\sys \  incorporates STGCN\cite{Yu_2018} to extract features from time-series graph data to consider the spatial and temporal relationship simultaneously.
To characterize the spatial correlation, cells are connected with the nearest six cells by edges.
The weight of the edge is a two-dimensional value.
The first weight is the geographical distance between the centroid of cells that is normalized by cell size.
The second weight of the edge is the distance along the road network. We calculate the shortest path along the road network by using the Open Street Map API. 
From the temporal view, we feed the recent 6-time slot historical demand graphs to the STGCN.

Although the user needs the recent historical taxi demand of the target regions due to this module, it can not be a heavy task because our framework requires only the recent 6-time slot historical demand records in the entire target region.
For example, crowd-sourcing can solve this problem easily.

\subsubsection{Mobility view}
\label{sec:mob-view}
While short-term demand trends are essential to predict next-timestep demand, mobility trends within the region, which are long-term tendencies, are also useful for demand prediction.
For example, taxi demand from the restaurant district to a residential area at night may be high compared to other origin-destination (OD) pairs.
Different from the graph of the spatio-temporal view, 
we first calculate the OD pair matrix which represents the number of demands from a \textit{cell} toward the drop-off \textit{cell}, and from each cell, we connect edges to the top five cells at most. This connectivity equates to the mobility tendency of the region.
This statistical mobility tendency is easy to access because some administrative agencies provide a survey of it, or the users can substitute the values they can get from map monitoring APIs.

% \subsubsection{Semantic view}
% \label{sec:semantic-view}
% Intuitively, cells with similar functionality in regions may have similar demand patterns.
% For example, residential areas may have a high number of demands in the morning when people commute, and commercial areas may expect to have high demands on weekends.
% Although a similar area may not necessarily be close geographically, this similarity is useful for predicting taxi demand.
% To represent this relationship, we construct a functional similarity graph named a semantic graph among regions.
% We set up the functionality of the cell(called "landuse") as a feature by using the APIs provided by Foursquare\footnote{https://foursquare.com/}.

\subsubsection{Semantic view}
\label{sec:semantic-view}
\color{black}
Intuitively, spatial cells with similar urban functionality tend to exhibit similar taxi demand patterns, even when they are not geographically adjacent.
For example, residential areas often show demand peaks during commuting hours, whereas commercial or entertainment districts typically experience higher demand during weekends or nighttime.
Such functional regularities provide information that is complementary to spatial proximity and mobility-based interactions, as they reflect demand behaviors driven by land-use characteristics rather than physical distance.

To capture these functional relationships, we construct a semantic view based on point-of-interest (POI) information.
In this view, the graph structure follows the same geographically defined topology used for modeling historical taxi demand, where nodes correspond to spatial cells and edges represent geographic adjacency.
The distinction from the mobility view lies not in the graph connectivity, but in the node attributes.

Specifically, POIs obtained from the Foursquare API\footnote{https://foursquare.com/} are aggregated within each cell according to predefined categories (e.g., residential, commercial, office, and entertainment), forming a semantic feature vector that characterizes the land-use composition of the cell.
These categorical POI features are then mapped into continuous representations through an embedding layer, enabling the model to learn compact and trainable semantic descriptors.
By propagating the embedded POI features over the geographically structured graph, the semantic view facilitates information sharing among neighboring cells with similar functional roles, allowing the model to capture recurring demand patterns associated with land-use similarity.
This design is particularly effective for taxi demand prediction, as functionally similar areas often exhibit correlated temporal demand behaviors even in the absence of strong mobility interactions.
\color{black}

\subsubsection{Network architecture}
\label{sec:network_arch}
\begin{figure}[tb]
%\vspace{-0.5cm}
    \centering
    \includegraphics[width=8cm]{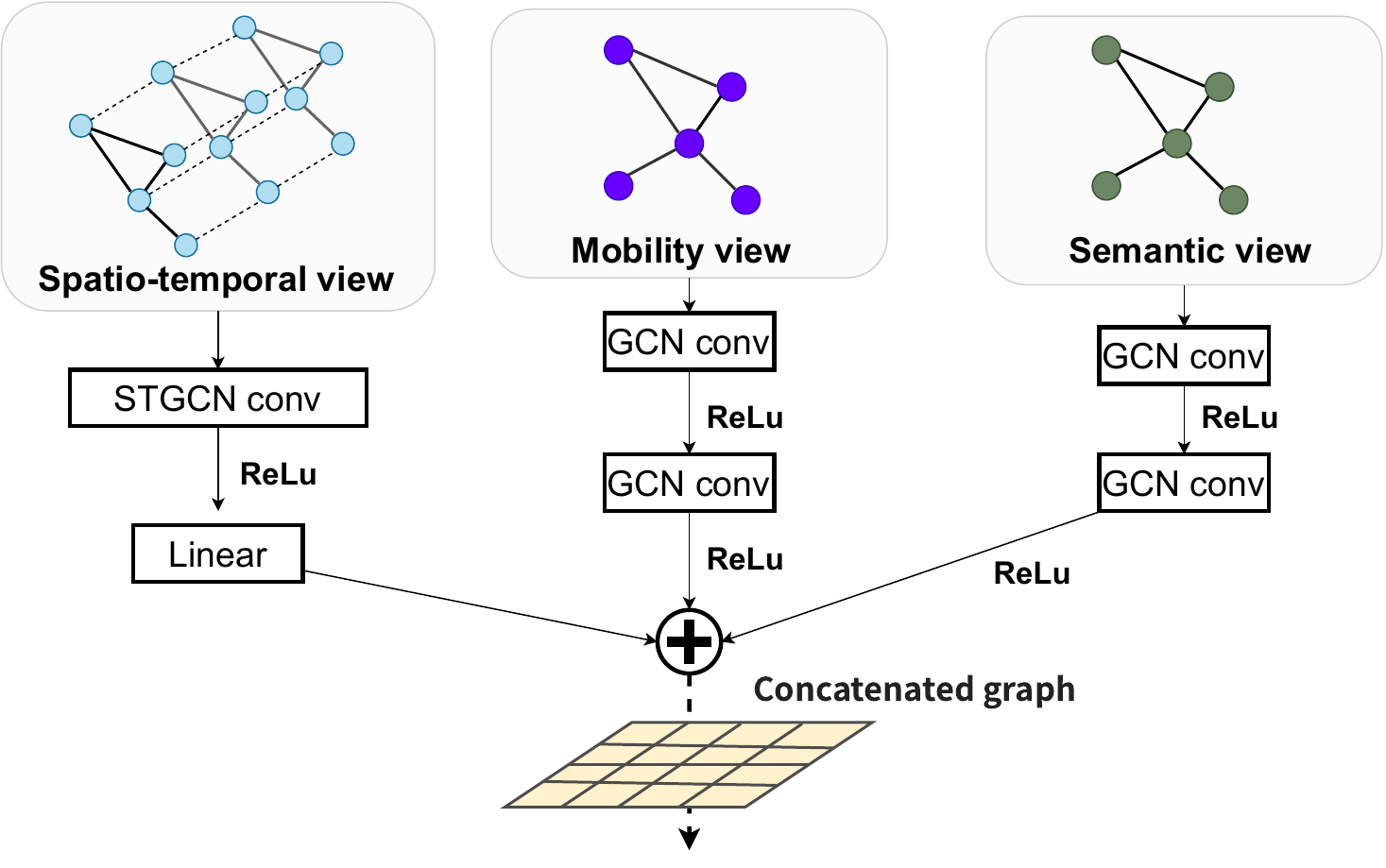}
    \caption{Graph processing module and network architecture.
    \sys\ employs three views of graph processing: Spatio-temporal view, Mobility view, and Semantic view.
    Three views of the graph are processed and concatenated by this module, and then fed into the region-agnostic module.}
    %%\vspace{-0.4cm}
    \label{fig:graph_processing_detail}
    % %\vspace{-0.6cm}
\end{figure}

Fig.~\ref{fig:graph_processing_detail} shows the overview of the graph processing module of \sys.
As we mentioned above, we employ three types of view graphs to predict taxi demand accurately.
To leverage each view graph, we design the graph convolution network for each view.
In order to treat the region as a graph $G$, we define the graph for each region as $G = (V,E)$, where the set of locations $L$ are nodes $V = |L|$, $E \in V \times V$ is the edge set.
The spatio-temporal graph processing network takes the most recent $h$ time steps historical taxi demand sequence $\{D_{t-h+1}, D_{t-h+2}, \dots, D_t\}$ as input to learn the historical spatial-temporal patterns.
These $h$ slots demands are combined and organized into a 3-D matrix, shape $h \times V \times d$, where $d$ is the node feature.
In this paper, since we only use demand as a node feature for the spatio-temporal view, the dimension of $d$ is 1.
We convert spatio-temporal graph (historical demand sequence) to feature representation (size is $\mathbb{R}^{V \times h_{ST}}$) by using STGCN\cite{Yu_2018}, where $h_{ST}$ is the number of hidden channels.
% STGCN\cite{Yu_2018} can receive the temporal graph $H \in \mathbb{R}^{M \times V \times C}$ where M is timestep, n is the number of nodes and C is the number of node feature.

Semantic and mobility view graphs are extracted features by GCN\cite{kipf2017semisupervised}.
The semantic graph processing network takes the static graph as a 2-D matrix, shape $V \times S_n$ to learn the relationship of functional similarity and demand pattern, where $S_n$ is the dimension of the semantic node feature.
The output channel of the semantic graph network is $h_{S}$.
The mobility view graph is processed in the same way as the mobility view. 
The output channel of the mobility graph network is $h_{M}$.
To treat three different views of the graph as features simultaneously, we concatenate these graphs and make a 2-D matrix, shape $V \times H$, where $H = h_{ST} + h_{S} + h_{M}$.

% \textcolor{blue}{Taxi-demand observations are spatially imbalanced over the hexagonal grid in some cases, where a small number of cells can be much denser than others.
% Our framework mitigates this issue in two ways.
% First, we include a semantic view as part of the input features in addition to the historical demand view.
% This provides complementary signals for sparsely sampled cells, so the model does not rely solely on uneven demand history.
% Second, to further prevent dominant nodes from disproportionately influencing message passing, we employ attention-based graph convolution and graph normalization in the graph encoder.
% Graph attention adaptively weights neighborhood contributions.}
\textcolor{black}{Taxi-demand observations can be spatially imbalanced over the hexagonal grid, where a small number of cells may contain substantially denser samples than others.
To mitigate this imbalance, our framework leverages both multi-view input features and a robust graph encoder.
Specifically, we incorporate a semantic view in addition to the historical demand view, providing complementary, non-count-based signals for sparsely sampled cells and reducing reliance on uneven demand histories.
Moreover, the graph encoder combines attention-based graph convolution with graph normalization to limit the influence of dominant nodes during message passing.
Graph attention adaptively weights neighborhood contributions, while normalization stabilizes node representations within each graph.}

\textcolor{black}{
Through the multi-view graph networks, the graph encoder learns latent node representations that capture complementary aspects of urban structure and dynamics, including functional similarity derived from the semantic view and aggregated movement patterns encoded in the mobility view.
However, under conventional supervised training, the encoder produces entangled latent representations in which region-agnostic relational patterns are mixed with region-dependent characteristics of the training data, such as absolute demand magnitude rather than temporal demand structure, dominant travel routines, and local regulations on taxi supply.
As a result, the model can achieve strong predictive performance within observed regions by predominantly relying on region-dependent characteristics, which in turn limits generalization to unseen regions.
To achieve region-agnostic demand representation, it is therefore necessary to extract region-agnostic features, corresponding to transferable relational patterns captured by the multi-view graph encoder.
This requirement motivates a learning strategy that separates region-agnostic feature from entangled latent representations from multi-view graph encoder, enabling the model to focus on generalizable urban dynamics across regions.
}

\subsection{Region-Agnostic Feature Extraction} \label{sec:region_neutral}

\color{black}
The region-agnostic feature extraction module plays a crucial role in extracting region-agnostic latent features by suppressing region-dependent components associated with taxi demand prediction.
The high coupling between the input data and its corresponding region poses a challenge for generalization.
Consequently, this network is trained to perform the separation task, facilitating the projection of input data into two distinct spaces: region-specific and region-agnostic.
\textcolor{black}{
Through this design, region-agnostic features are extracted from entangled latent representations derived from multi-view graphs.
This enables the model to maintain accurate taxi demand prediction performance even in previously unseen regions, as illustrated in Fig.~\ref{fig:feature_extraction_with_geo_visualization}.
}

\subsubsection{Network Architecture}
\label{sec:region-neut-network-arch}
\begin{figure}[tb]
    \centering
    \includegraphics[width=8cm]{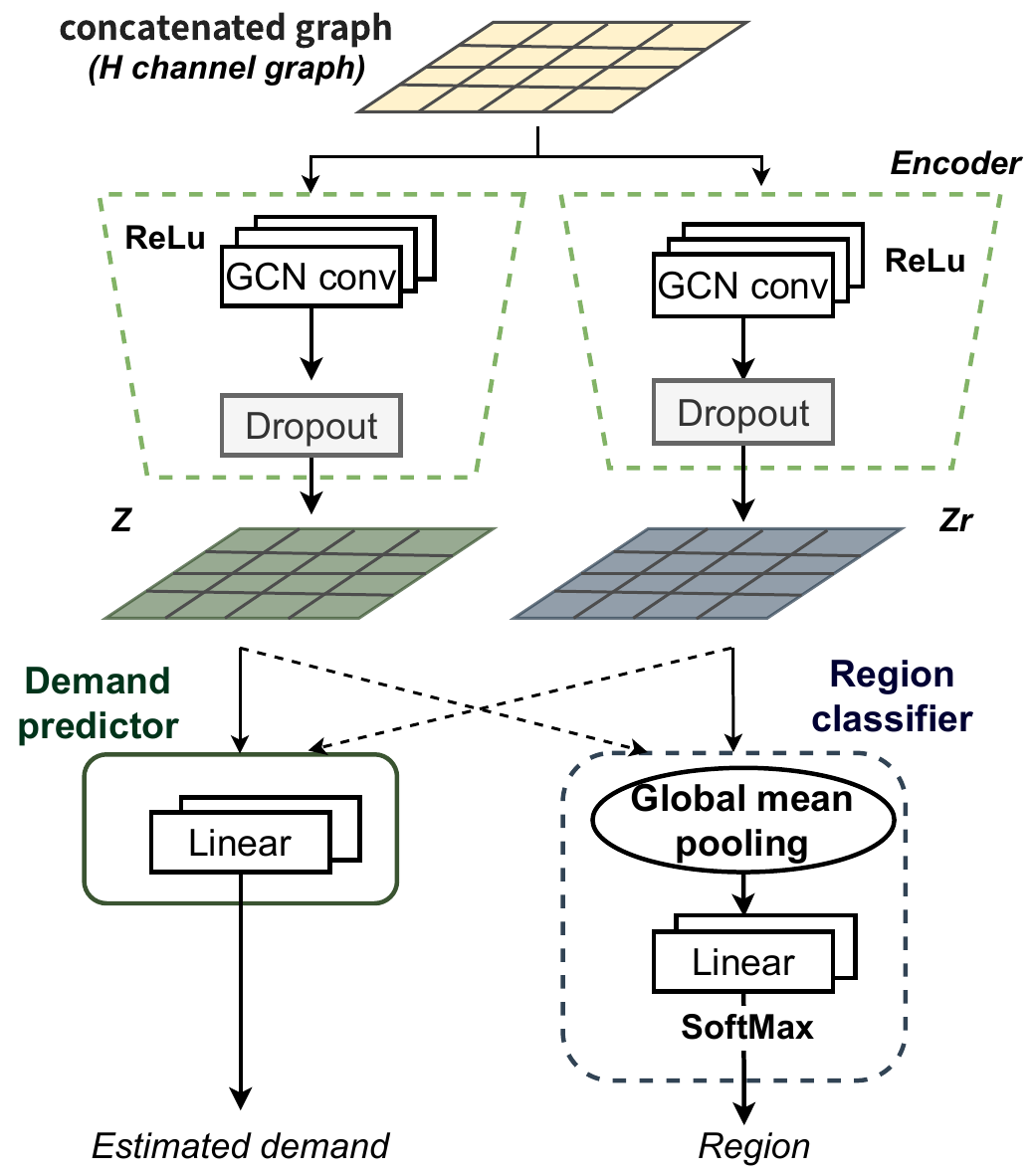}
    \caption{Region agnostic module network architecture.
    In this figure, $z$ and $z_r$ represent region-agnostic features and region-specific features,  respectively.
    The demand prediction module is responsible for predicting taxi demand, while the region classifier determines the region to which a given latent feature belongs.
    }
    \label{fig:region-neutral_architecture-region_neutral_mechanism}
    %\vspace{-0.6cm}
\end{figure}

The network architecture of the proposed \textcolor{black}{region-agnostic} approach is shown in Fig.~\ref{fig:region-neutral_architecture-region_neutral_mechanism}. It comprises five sub-networks: two encoders, a decoder, a demand prediction module, and a region classifier.
To extract latent features that capture the \textcolor{black}{region-agnostic} and region-specific factors, we utilize two encoders: $q(z|x^{r})$ and $q_r(z_r|x^{r})$ parameterized by $\Phi$ and $\Phi_r$, respectively. 
Here, $x^{r}$ represents the input feature. Both encoders share the same structure, consisting of three Graph Convolutional Network (GCN) convolutional layers and ReLU activation functions, as shown in Fig.~\ref{fig:region-neutral_architecture-region_neutral_mechanism}.
\textcolor{black}{These two encoders serve as variational inference networks that parameterize two separate approximate posterior distributions for $z$ and $z_r$ from the same input $x^{r}$, thereby enabling factorized latent representations for region-agnostic demand prediction and region-specific characteristics.}
The decoder, denoted as $p(x^{r}|z_r,z;\phi)$, aims to reconstruct the input as $\hat{x}^{r}$ using the extracted latent features $z$ and $z_r$. It is constructed using the InnerProductDecoder \cite{kipf2016variational}.
The demand prediction module $SR_y$ is responsible for predicting taxi demand, while the region classifier $SC_r$ determines the region to which a given latent feature belongs. Both modules are implemented as fully connected (FC) neural networks. As the region classifier performs graph classification, a global mean pooling layer is applied before the FC model.

\subsubsection{Loss function}
\label{sec:loss_function}
To effectively extract both \textcolor{black}{region-agnostic} and region-specific factors, we employ three loss functions: Variational Graph Autoencoder (VGAE) loss ($L_{elbo}$), task-specific loss ($L_{TS}$), and Independent Excitation loss ($L_{IE}$).
\textcolor{black}{The VGAE loss corresponds to an evidence lower bound objective, where the decoder reconstructs $x^{r}$ from the latent variables $(z, z_r)$ while the KL terms regularize the approximate posteriors toward their priors.}
The VGAE architecture in \cite{kipf2016variational}, is utilized to reduce the dimensionality of feature graphs without losing important information. The learning process involves training probabilistic encoders and a decoder, which is achieved through the following loss function eq.\ref{eq:loss_function_elbo}:
\begin{equation}
\begin{split}
\label{eq:loss_function_elbo}
    L_{elbo} &= E_{r,q_{r}(z_r|x^{r}; \Phi_{r}), q(z|x^{r}; \Phi)}[\log p(x^{r}|z_r,z;\phi)] \\
        &-KL(q_r(z_r|x_r;\Phi_r)||p(z_r)) \\
        &-KL(q(z|x_r;\Phi)||p(z))
\end{split}
\end{equation}

where the first term is the reconstruction error, which measures the deviation between the original features $x^{r}$ and the reconstructed features $p(x^{r}|z_r,z;\phi)$. The last two terms calculate Kullback-Leibler (KL) divergence between the sampled latent features and corresponding priors, which are interpreted as regularizers on the latent feature spaces.
Since we adopt the VGAE architecture, this learning procedure is in a self-supervised manner.
where $KL[q(\cdot)||p(\cdot)]$ is the KL divergence between $q(\cdot)$ and $p(\cdot)$. 

The prior distribution $p(z)$ is assumed to be a Gaussian distribution $N(\cdot)$ with zero mean and unit variance for each dimension, as:
\begin{equation}
\label{eq:prior_distribution}
    p(z) = \prod_i p(z_i) = \prod_i N(z_i | 0, \textbf{I})
\end{equation}

To guide the feature learning process, we introduce two tasks by two classifiers: the demand prediction regressor $SR_y$ and the region classifier $SC_r$ with parameters $w_y$ and $w_r$, respectively. These modules take the latent features obtained from their respective encoders as input and predict their corresponding labels. Specifically, $SC_r$ classifies region label $r$ from the region-specific features $z_r$, while $SR_y$ predicts demand from the region-agnostic features $z$.
\textcolor{black}{This design explicitly assigns different predictive roles to $z$ and $z_r$, which complements the reconstruction objective in $L_{elbo}$ and promotes task-aligned factorization of the latent space.}

To achieve task-specific representations (i.e., two representations, one for region-specific information and another for region-agnostic demand prediction), we define the task-specific loss function ($L_{TS}$) as: 
\begin{equation}
\begin{split}
\label{eq:loss_function_ts}
    L_{TS} = \frac{1}{N_s}\sum_{r=1}^{N} \sum^{N_r}_{i=1}[&MAE(y_i^{r}, SR_y(z; w_y)) \\
&+ \ell(r, SC_r(z_r; w_r))]
\end{split}
\end{equation}

This function, expressed in eq.\ref{eq:loss_function_ts}, calculates the average loss over all regions and data points using categorical cross-entropy $\ell(\cdot)$ and mean absolute error $MAE(\cdot)$ where $N_r$ is the number of data for region $r$.
The objective of this loss function is to optimize the classifiers $SC_r$ and regressor $SR_y$ to accurately predict their respective labels. Minimizing this loss plays a critical role in facilitating the disentanglement of different task-specific information (encompassing both region-specific details and region-agnostic demand information) within the latent features. 

To further ensure the separation of the input into distinct encoded representations of non-overlapping information, we incorporate an Independent Excitation mechanism. This mechanism is designed to minimize the accuracy of the classifier $SC_r$ when the feature $z$ is fed into it, as follows:
% \color{blue}
\begin{equation}
\begin{split}
\label{eq:loss_function_ie}
    L_{IE} = -\frac{1}{N_s}\sum_{r=1}^{N} \sum^{N_r}_{i=1}[&MAE(y_i^{r}, SR_y(z_r; w_y)) \\
+&\ell(r, SC_r(z; w_r))]
\end{split}
\end{equation}
Where $N_s=\sum_{i=1}^{r} N_r$. It is noteworthy that the minimization of eq.\ref{eq:loss_function_ts} and eq.\ref{eq:loss_function_ie} leads to latent representations $z$ and $z_r$ becoming more representative of each respective task while being less representative for another task. Specifically, $z$ is designed to be highly informative for predicting taxi demand, while its capability for distinguishing regions is intentionally diminished. As such, $z$ serves as a \textcolor{black}{region-agnostic} representation specifically tailored for taxi demand prediction. Moreover, when eq.\ref{eq:loss_function_elbo} is minimized, both $z$ and $z_r$ can be effectively utilized for reconstructing the original information. Consequently, $z$ and $z_r$ not only function as independent features but also encompass complete information necessary for reconstruction, incorporating the reconstruction loss outlined in eq.~\ref{eq:loss_function_elbo}.

Finally, the overall loss function of the model is expressed as follows:
\begin{equation}
\begin{split}
\label{eq:loss_function}
    L = L_{elbo} + \alpha L_{TS} + \beta L_{IE}
\end{split}
\end{equation}
Where $\alpha$ and $\beta$ are hyperparameters. Notably, larger values of $\alpha$ result in latent features $z$ and $z_r$ attaining higher accuracy for their respective tasks, yet potentially compromising the independence between them. Careful consideration and tuning of $\alpha$ and $\beta$ are crucial for achieving an optimal balance between task accuracy and feature independence within the model.
\textcolor{black}{During inference, we use the region-agnostic latent feature $z$ as the input to the demand prediction module $SR_y$, which enables deployment to previously unseen regions.}

\subsection{Online Demand Prediction}\label{sec:online_dp}
During the online stage, an end user, such as a taxi service provider, can utilize the framework to forecast the unknown demand pattern in any arbitrary region, including previously unobserved regions. This predictive capability is achieved through a series of systematic steps. Firstly, the framework prepares and processes the associated views using a graph processing module, ensuring that all relevant data of the region under consideration is considered. These processed views are then concatenated, forming a comprehensive multi-view graph structure that represents the region.  Subsequently, the derived multi-view graph is input into a pre-trained encoder model to extract \textcolor{black}{region-agnostic} features. Finally, the trained taxi demand prediction model utilizes these \textcolor{black}{region-agnostic} features to forecast the taxi demand within that target region. This predictive capability provides valuable insights to the end user, aiding in decision-making processes and resource allocation.

\color{black}

\section{Evaluation} \label{sec:evaluation}

\color{black}
\subsection{Data Collection and Configuration}\label{sec:data_collection}
In this section, we discuss the datasets used to evaluate the proposed approach, including our proprietary dataset and publicly available datasets.

\subsubsection{Our dataset}
\label{sec:our-dataset}

% \textcolor{blue}{
% % In this study, we collected real-world taxi operation data for over one year (2021-10-29 to 2023-04-27) from 20 taxi service providers in Japan.
% The dataset consists of GPS-based taxi trip records, where each trip is annotated with pick-up and drop-off event labels, second-level timestamps, and corresponding latitude and longitude coordinates.
% In total, more than 110,000 taxi trips were recorded.
% The details of our dataset are summarized in Table \ref{tab:ourdataset}.
% To ensure validity and comprehensiveness in our experiments, we have chosen the five datasets (representing distinct cities or regions) that boast the highest volume of data recorded by the taxi service providers.
% }
\textcolor{black}{
In this study, we collected real-world taxi operation data for over one year (from October 29th, 2021, to April 27th, 2023) from twenty taxi service providers across five distinct regions in Japan.
The dataset consists of GPS-based taxi trip records, where each trip is annotated with pick-up and drop-off event labels, second-level timestamps, and corresponding latitude and longitude coordinates.
In total, more than 110,000 taxi trips were recorded.
Each region corresponds to the service area covered by one or more taxi providers, and these regions do not geographically overlap with one another.
The details of our dataset are summarized in Table~\ref{tab:ourdataset}.
}

\begin{table*}[tb]
    \centering
    \caption{Our Dataset description.}
    \label{tab:ourdataset}
    \begin{tabular}{c||ccp{1.2cm}} \hline
        Dataset & Duration & Area[km $\times$ km] & \#records \\ \hline
        \textbf{r1} & 2021-10-30 $\sim$ 2022-03-19 \quad (5 months) & $36.6 \times 38.6$ & 40.3k \\ 
        \textbf{r2} & 2022-02-11 $\sim$ 2023-04-27 \quad (14 months) & $16.6 \times 13.5$ & 39.3k \\ 
        \textbf{r3} & 2021-10-30 $\sim$ 2022-03-18 \quad (5 months) & $14.4 \times 14.4$ & 6.9k \\
        \textbf{r4} & 2021-10-28 $\sim$ 2022-03-26 \quad (5 months) & $31.1 \times 31.5$ & 15.4k \\ 
        \textbf{r5} & 2021-10-29 $\sim$ 2022-03-28 \quad (5 months) & $16.6 \times 11.7$ & 6.7k \\ \hline
    \end{tabular}
\end{table*}

\begin{table*}[tb]
    \centering
    \caption{Open Dataset description.}
    \label{tab:opendataset}
    \begin{tabular}{c||ccp{1.2cm}} \hline
        Dataset &  Duration & Area[km $\times$ km] & \#records \\ \hline
        \textbf{NYC yellow taxi trip data} & 2019-01-01 $\sim$ 2019-01-31 \quad (1 month) & $39.9 \times 41.9$ & 100k\\ 
        \textbf{Taxi trips reported to the City of Chicago} & 2022-01-01 $\sim$ 2022-01-31 \quad (1 month) & $37.1 \times 29.1$ & 308k \\
        \textbf{Mobility traces of taxi cabs in San Francisco, USA} & 2008-05-17 $\sim$ 2008-06-10 \quad (1 month)& $77.2 \times 26.3$ & 850k\\ \hline
    \end{tabular}
\end{table*}

\subsubsection{Open dataset}
\label{sec:open-dataset}
\sys \  is evaluated using publicly available datasets for the purpose of benchmarking, as described below. We make use of open data collected from various cities in the United States, with each city considered as a distinct region. While these datasets may contain additional features for other applications, we have extracted the specific attributes required for our taxi demand application, namely latitude, longitude, and timestamp indicating the pick-up or drop-off time of each taxi passenger. 
% Additionally, we enhance the datasets by appending the "landuse" data to each demand. This additional information provides insights into the land usage patterns associated with each taxi demand.

Table \ref{tab:opendataset} provides a detailed description of the open dataset.
% The first dataset we utilize is the \textbf{NYC Yellow Taxi Trip Data}\footnote{https://www.nyc.gov/site/tlc/about/tlc-trip-record-data.page}. This dataset comprises trip data for Yellow Taxis in New York City.
% The second dataset is \textbf{Taxi Trips Reported to the City of Chicago}\footnote{https://data.cityofchicago.org/Transportation/Taxi-Trips/wrvz-psew}, which covers the Chicago region.
% The last dataset represents the \textbf{Mobility Traces of Taxi Cabs in San Francisco, USA} \cite{c7j010-22}. This dataset also includes latitude, longitude, and timestamp information, with a temporal granularity of one second.
\textcolor{black}{
The first dataset is the \textbf{NYC Yellow Taxi Trip Dataset}%
\footnote{\url{https://www.nyc.gov/site/tlc/about/tlc-trip-record-data.page}}.
This dataset contains trip records of New York City yellow taxis.
In this dataset, pick-up and drop-off locations are encoded using predefined zone identifiers rather than raw latitude and longitude coordinates.
To obtain spatial coordinates, we assign each record the geometric center (centroid) of the corresponding zone.
The data were collected from January 1, 2019 to January 31, 2019, comprising approximately 100K trip records.
The second dataset is the \textbf{Taxi Trips Reported to the City of Chicago}%
\footnote{\url{https://data.cityofchicago.org/Transportation/Taxi-Trips/wrvz-psew}}.
This dataset covers taxi trips within the Chicago metropolitan area.
Unlike the NYC dataset, it directly provides latitude and longitude coordinates, along with timestamps rounded to the nearest 15 minutes.
The data collection period spans from January 1, 2022 to January 31, 2022, and includes approximately 308K trip records.
The third dataset is the \textbf{Mobility Traces of Taxi Cabs in San Francisco, USA} \cite{c7j010-22}.
This dataset also includes precise latitude and longitude coordinates, with timestamps recorded at a one-second granularity.
The data were collected between May 17, 2008, and June 10, 2008.
The dataset contains approximately 850K records.
}

% \textcolor{blue}{
% To encode locations, we convert absolute latitude and longitude values into relative latitude/longitude values. Specifically, we define the center of the target area as a reference point and represent each location by the relative differences from this center, i.e., ($\Delta$lat, $\Delta$lon). This conversion into relative latitude/longitude ensures the generalization ability of the model and the efficiency of learning.
% To calculate demand, we apply virtual gridding to the target area.
% We overlay a virtual grid and assign each pick-up and drop-off event to a grid cell according to its location.
% We then count the number of events in each cell to construct a demand map, where each grid cell stores an integer demand value.
% }

\color{black}
\subsubsection{Common preprocessing and demand construction}
\textcolor{black}{
For both datasets, we apply the same preprocessing pipeline.
To encode locations, we convert absolute latitude and longitude values into relative latitude/longitude values.
Specifically, we define the center of the target area as a reference point and represent each location by the relative differences from this center, i.e., ($\Delta$lat, $\Delta$lon). This conversion into relative latitude/longitude ensures the generalization ability of the model and the efficiency of learning.
To construct taxi demand, we first apply a virtual grid to the target area and assign each pick-up and drop-off event to a corresponding grid cell based on its geographic location.
For each grid cell and each time interval, we count the number of assigned events to compute the cell-level demand.
In the temporal dimension, as summarized in Table~\ref{tab:model_parameter}, time is discretized into 30-minute intervals.
% Specifically, for a given grid cell, all pick-up and drop-off events occurring within the same 30-minute interval are summed, and the resulting count is computed as the taxi demand of that cell at that time.
Specifically, within each 30-minute interval, we sum pick-up and drop-off events in each grid cell and use the total as the taxi demand.
Additionally, we augment each grid-cell demand with the corresponding "landuse" information, which provides contextual cues about the local land-use patterns associated with the demand.
}
% For this study, we only use the pick-up and drop-off points to calculate the demand. The total number of records in our dataset is over 130,000. The details of our dataset are summarized in Table \ref{tab:ourdataset}. To ensure validity and comprehensiveness in our experiments, we have chosen the five datasets (representing distinct cities or regions) that boast the highest volume of data recorded by the taxi service providers.

\color{black}

\begin{table}[!t]
    \centering
    %\vspace{-0.1cm}
      \caption{Default system parameters.}
      %\vspace{-0.1cm}
    \resizebox{\linewidth}{!}{%
    \begin{tabular}{c|c|l}
    \hline
        Parameter & Default value & Explored range \\
    \hline
        Learning rate &  $10^{-4}$ & $\{10^{-3}, 10^{-4}, 10^{-5}, 10^{-6}\}$\\
        Batch size & 64 & $\{16, 32, 64, 128\}$ \\
        Optimizer & AdamW & -- \\
        Max epoch number & 50 & $\{5, 10, 50, 100\}$ \\
        Hidden channel of $z$\&$z_r$ & 32 & $\{16, 32, 64, 128\}$ \\
        $\alpha$ & 10 & $\{0.01, 0.1, 1, 10, 100 \}$ \\
        $\beta$ & 1 & $\{0.01, 0.1, 1, 10, 100 \}$ \\
        Hexagon cell edge [km] & 1.4 & $\{1.4, 3.7, 9.9 \}$ \\
        Time Interval [minute] & 30 & $\{10, 30, 60 \}$ \\

    \hline
    \end{tabular}
    }
  \label{tab:model_parameter}
    %\vspace{-0.5cm}
\end{table}

\subsection{Evaluation Metrics}
\label{sec:evaluation-metrics}
% We use classification accuracy to evaluate the proposed method for our dataset and open data.
% When we use our dataset, we adopt three classes for demand prediction, that is low, mid, and high.
% \hl{W1-accuracy -> changeable }
To assess our proposed method's performance, we calculate the prediction "accuracy" accounting for one-class error. A prediction is considered correct if the absolute error between the predicted and actual taxi demand is less than two. This aligns with real-world taxi dispatching systems where estimating demand within a certain margin of error is acceptable\cite{miao2017data, miao2015robust}.
% \hl{[do you have reference for that <= I add reference(by ozeki)]} 
We evaluate the region-agnostic approach by measuring the "accuracy" of the taxi demand prediction model in an unseen region. We employ a leave-one-city-out strategy for evaluating the models in both our dataset and the open dataset.

% Specially, we select one city as an unseen city from one dataset and apply the model trained by the rest region's data to the unseen region.

\color{black}

\subsection{Analysis of System Parameters and Modules}
\label{sec:analysis-of-sys}

In this section, we study the impact of the main system parameters and modules.
The default parameters are reported in Table \ref{tab:model_parameter}.

\subsubsection{Impact of Loss Function Weights}
\label{sec:impact-loss-func}
\begin{figure}[!t]
    \centering
    \includegraphics[width=8cm]{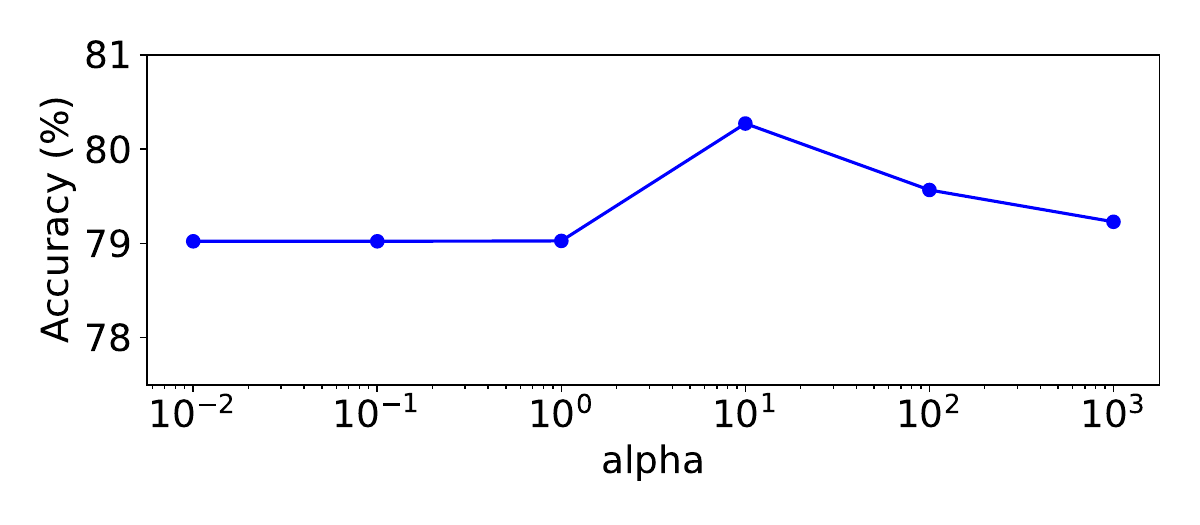}
    %\vspace{-0.5cm}
    \caption{The accuracy for various $\alpha$ ($\beta=1$).}
    \label{fig:performance_hyper-parameter_alpha}
    %\vspace{-0.5cm}
\end{figure}

\begin{figure}[!t]
    \centering
    \includegraphics[width=8cm]{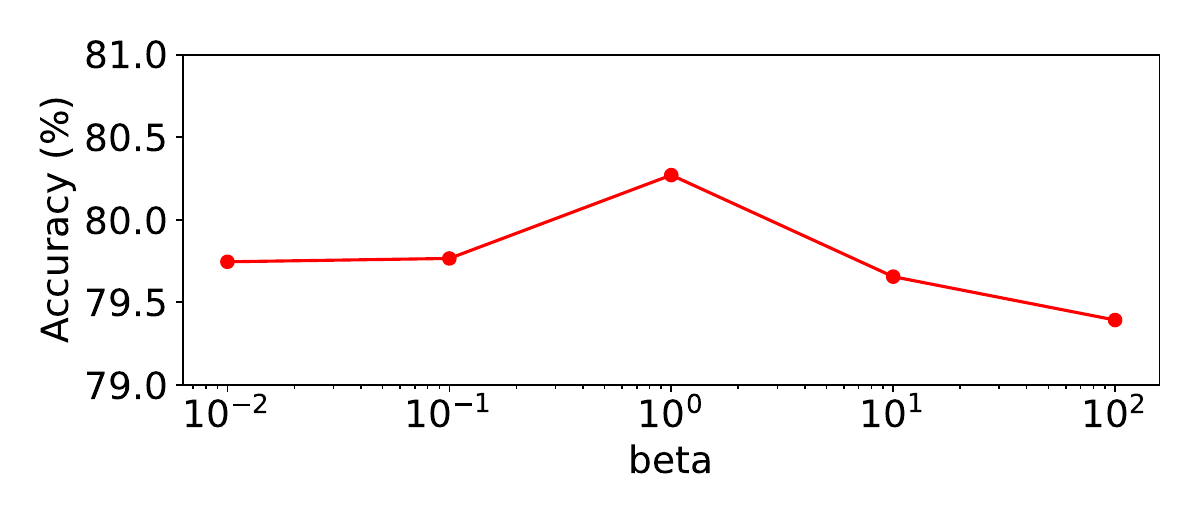}
    %\vspace{-0.5cm}
    \caption{The accuracy for various $\beta$ ($\alpha=10$).}
    \label{fig:performance_hyper-parameter_beta}
    %\vspace{-0.5cm}
\end{figure}

In this section, we examine the crucial role of the weights $\alpha$ and $\beta$ of the loss functions for achieving an optimal balance between task accuracy (taxi demand prediction and region discrimination) and the creation of a region-agnostic general representation.
Fig.~\ref{fig:performance_hyper-parameter_alpha} illustrates the model accuracy in unseen regions for various values of $\alpha$, while keeping $\beta$ fixed. Similarly, Fig.~\ref{fig:performance_hyper-parameter_beta} demonstrates the model accuracy in unseen regions for different values of $\beta$, with $\alpha$ held constant.
The results depicted in Fig.~\ref{fig:performance_hyper-parameter_alpha} and Fig.~\ref{fig:performance_hyper-parameter_beta} highlight the sensitivity of the model's performance to changes in these parameters. This sensitivity emphasizes the importance of carefully selecting the values of $\alpha$ and $\beta$ to achieve optimal outcomes.
The figures show that a balanced combination of learning the main task (demand prediction) and attaining a general \textcolor{black}{region-agnostic} representation occurs when setting $\alpha=10$ and $\beta=1$. This particular configuration leads to high performance in unseen regions, suggesting an effective trade-off between task accuracy and region-agnostic representation.

\subsubsection{Impact of Multi-view Graphs}\label{sec:multi_view_impact}
\begin{figure}[!t]
    \centering
    \includegraphics[width=8cm]{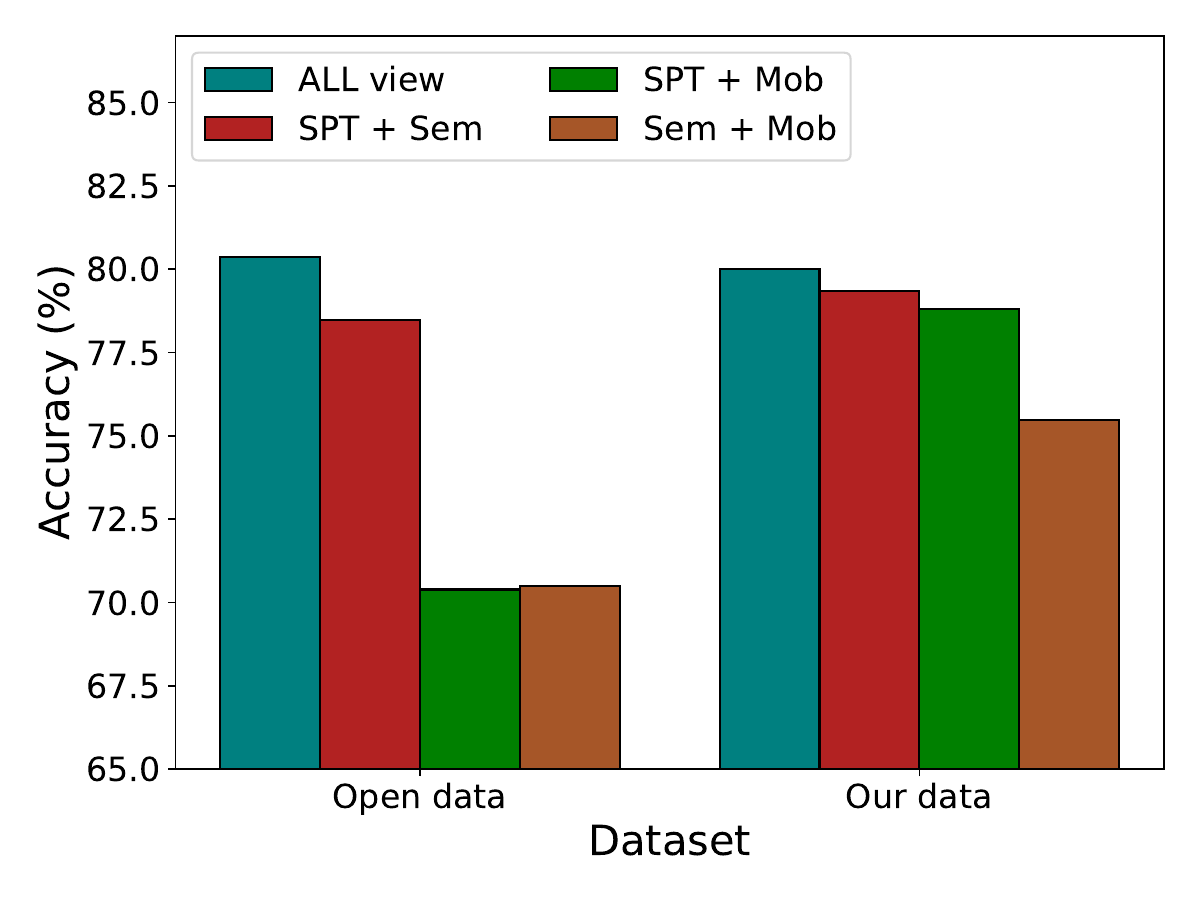}
    %\vspace{-0.5cm}
    \caption{Efficacy of multi-view graphs.  SPT, Sem, and Mob denote the Spatio-temporal graph, Semantic graph, and Mobility graph, respectively.}
    \label{fig:multi-view-effect}
    %\vspace{-0.5cm}
\end{figure}

In this section, we study the impact of collectively learning representations from multi-view graphs. Fig.~\ref{fig:multi-view-effect} presents the comparative results for each view of the graph, where SPT, Sem, and Mob denote the Spatio-temporal graph, Semantic graph, and Mobility graph, respectively.
The figure demonstrates that utilizing all three views yields superior performance compared to individual views or any combination of two views.
By integrating all three views, the model is able to leverage the complementary information present in each view leading to improved accuracy in prediction tasks.
Furthermore, the combination of Spatio-temporal and Semantic views exhibits consistently high accuracy. This can be attributed to the fact that these two views capture both the temporal dynamics and semantic context, allowing for a more comprehensive representation of the underlying data.
In contrast, the combination of Semantic and Mobility views shows a lower accuracy of 69.5\%. This result highlights the effectiveness of our Spatio-temporal view, as it contributes crucial information that is not adequately captured by the Semantic and Mobility views alone.

\color{black}
\subsubsection{Impact of Hexagonal Virtual Gridding}
\label{sec:impact-hex}
\begin{figure}[tb]
    \centering
    \includegraphics[width=8cm]{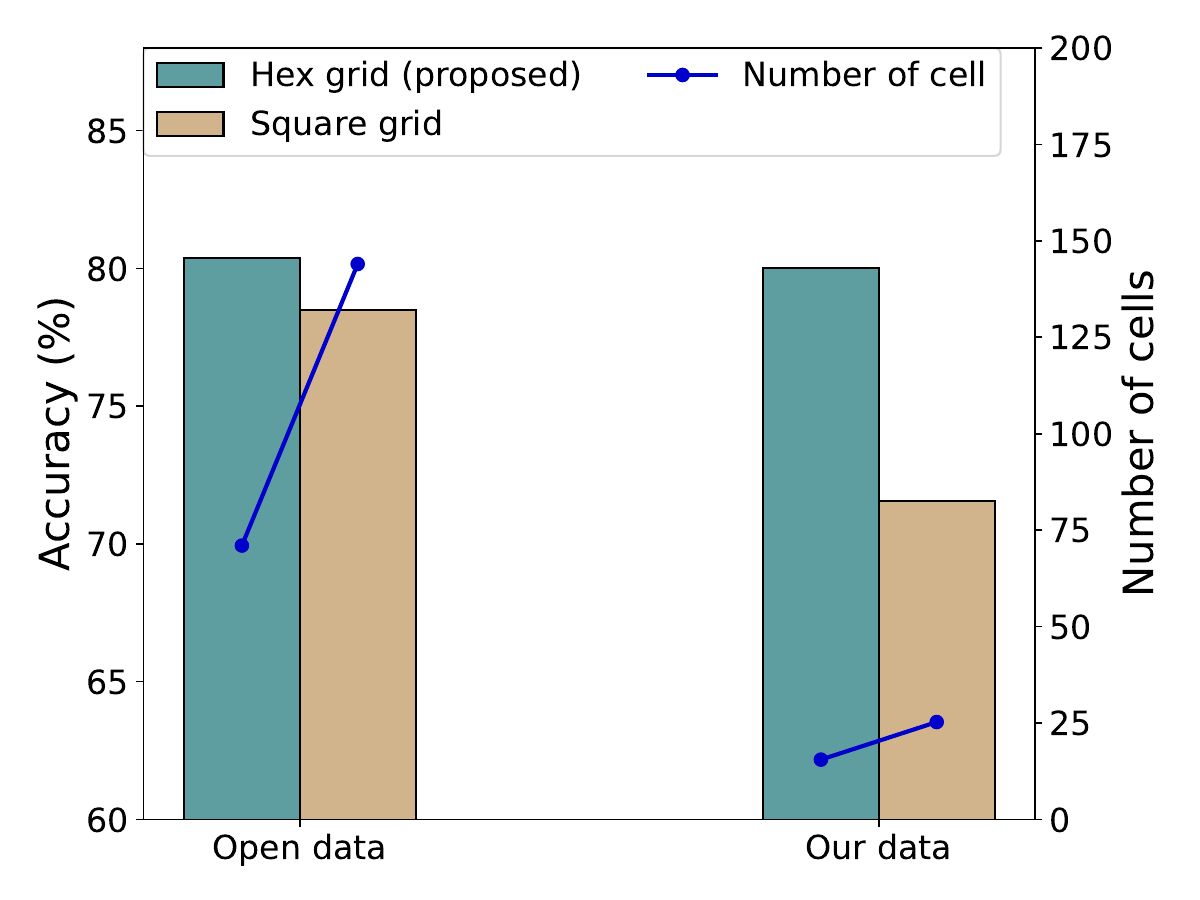}
    %\vspace{-0.5cm}
    \caption{Comparison with hexagonal and square gridding.}
    \label{fig:griding_system_w1-accuracy}
    %\vspace{-0.5cm}
\end{figure}

Fig.~\ref{fig:griding_system_w1-accuracy} presents a comparative analysis of the performance of our proposed taxi demand prediction approach, utilizing hexagonal and square grid structures, applied to both our proprietary and open data sources. The results demonstrate the superior effectiveness of the hexagonal grid structure in both datasets.
This superiority can be attributed to the unique properties of the hexagonal grid, which provides a more advanced representation of geographical features compared to its square counterpart. The hexagonal grid cells ensure a balanced neighborhood environment as each hexagon shares a border with six other hexagons. This characteristic promotes a more equitable distribution of cells, effectively mitigates edge effects, and accurately depicts spatial relationships.
Moreover, hexagonal cells offer the advantage of compact packing, allowing them to cover larger geographic areas with fewer cells. This quality refines the geographic representation while simultaneously reducing the grid's redundancy. Furthermore, due to their equidistant centers, hexagonal cells ensure consistent and regular spacing across the grid. This consistency facilitates precise distance calculations, enabling more robust spatial analysis and leading to improved accuracy in taxi demand prediction.

% \subsection{Flexibility in grid cell size}

% \begin{table}[tb]
%     \centering
%     \begin{tabular}{c|cccc}
%         \hline 
%        source res, target res  & (5,5) & (5,6) & (6,5) & (6,6) \\
%        \hline 
%        \hline 
%        W1-accuracy (our data) & xxx & xxx & xxx & xxx \\
%        W1-accuracy (open data) & 50.2 & 60.3 & 52.3 & 61.3 \\
%        \hline 
%     \end{tabular}
%     \caption{Cell size flexibility of proposed framework}
%     \label{tab:cell_size_flex}
% \end{table}

\subsection{Robustness Evaluation}
\label{sec:robust_eval}
% In this section, we evaluate the performance of the proposed approach when trained by target region data, which we call \textit{original model}.
% We measure the accuracy of the original model by splitting the target region's demand into training data and test data.
% Since the original model learns both region-specific and region-independent knowledge simultaneously, the original model specified this region.
% Fig.\ref{fig:accuracy_baseline} show the result of our proposed method and original model.
% Although the prediction accuracy of the original model is higher than the proposed method, the difference between the proposed method and the original model is less than $1\%$.
\begin{figure}[tb]
    \centering
    \includegraphics[width=8cm]{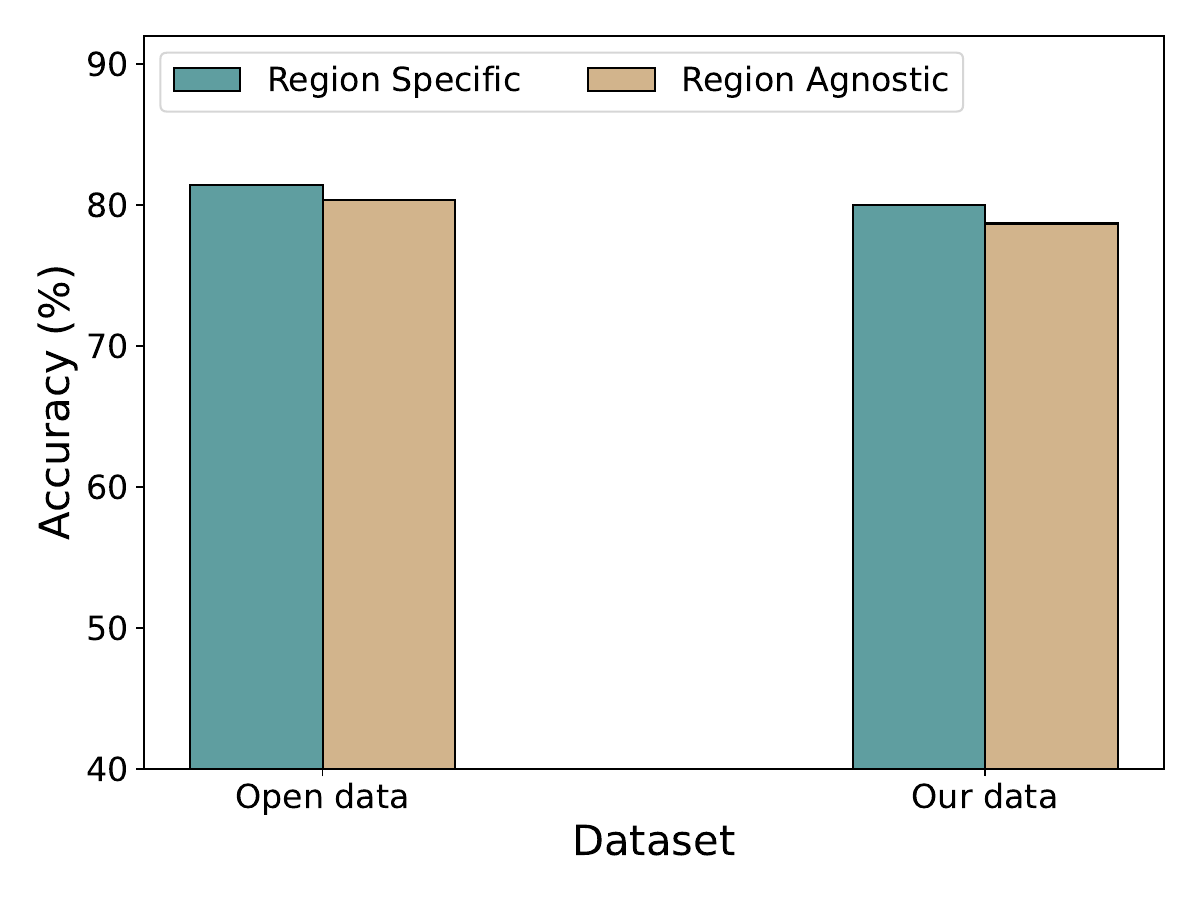}
    %\vspace{-0.8cm}
    \caption{Accuracy of region-specific and region-agnostic model. 
    The accuracy of a Region-specific model is the outcome of training and testing a single model using an 80-20 strategy within the same region.
    The accuracy of a Region-agnostic model is the outcome of training using leave-one-city-out and testing in an unseen region.
    }
    \label{fig:accuracy_baseline}
\end{figure}

\begin{figure}[tb]
    \centering
    \includegraphics[width=8cm]{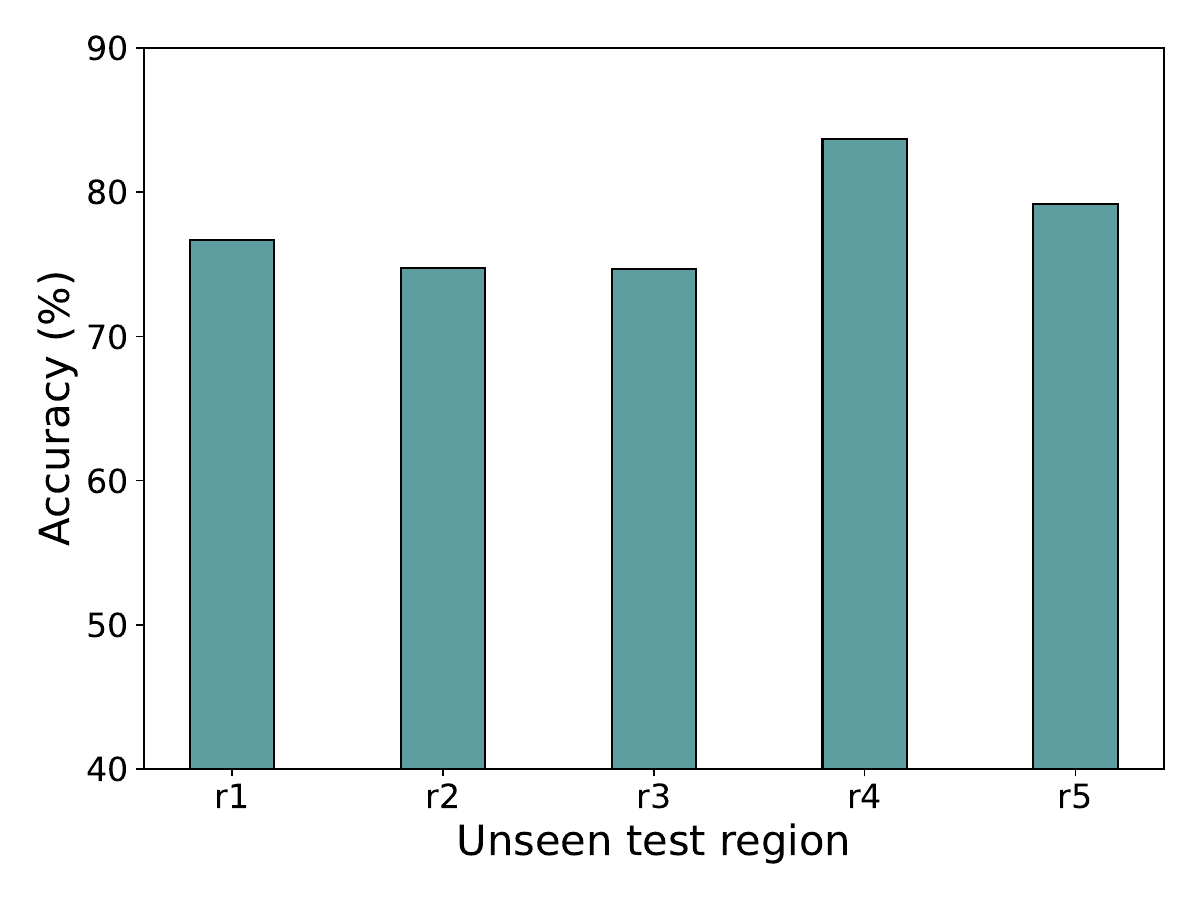}
    %\vspace{-0.8cm}
    \caption{Accuracy of each region (our dataset).}
    \label{fig:accuracy_each_region}
\end{figure}

\begin{figure}[tb]
    \centering
    \includegraphics[width=8cm]{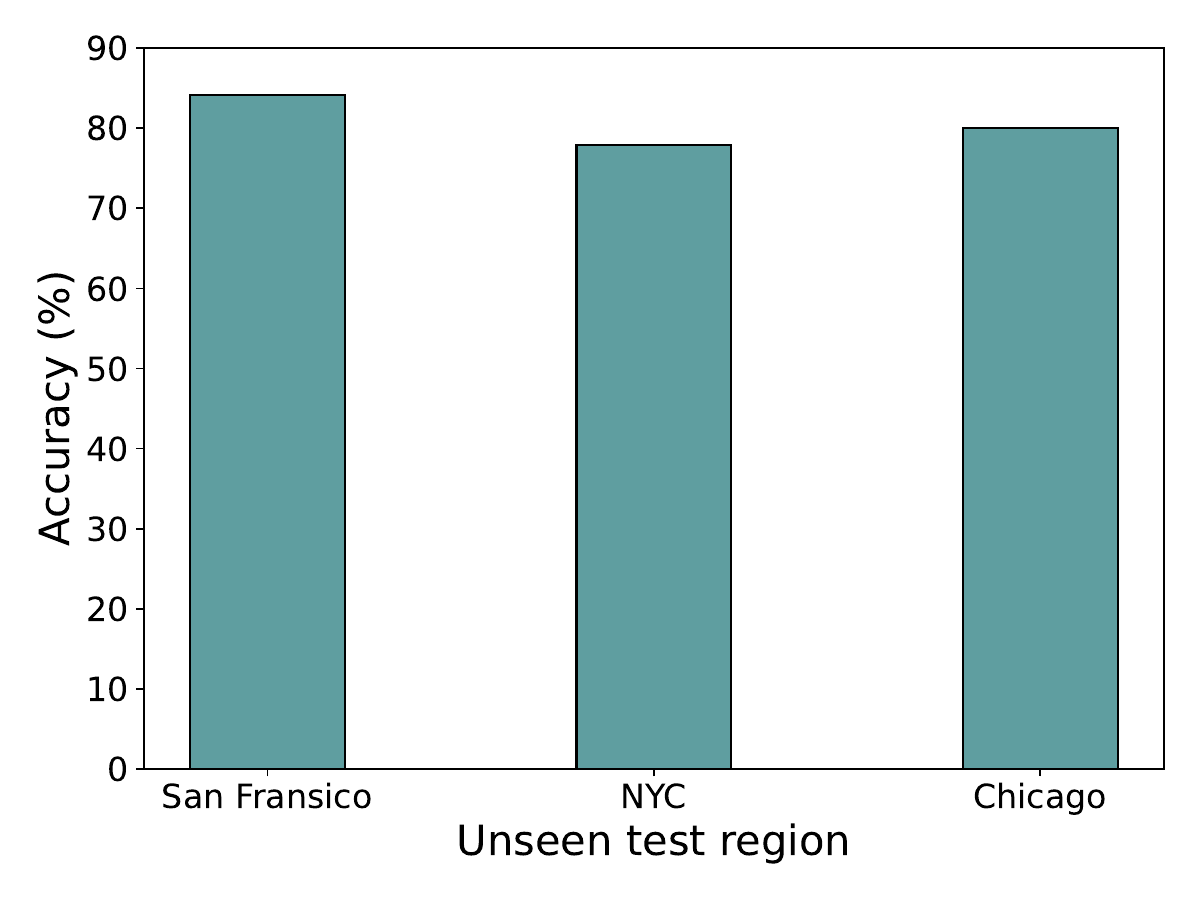}
    %\vspace{-0.8cm}
    \caption{Accuracy of each region (open dataset).}
    \label{fig:two_class_miss_each_region}
\end{figure}

% \begin{figure*}[tb]
%   \begin{minipage}[t]{0.31\linewidth}
%     \centering
%     \includegraphics[width=5.5cm]{imgs/accuracy_baseline.pdf}
%     %\vspace{-0.8cm}
%     \caption{Accuracy of region-specific and region-agnostic model}
%     \label{fig:accuracy_baseline}
%     %\vspace{-0.3cm}
%   \end{minipage}
%   \begin{minipage}[t]{0.31\linewidth}
%     \centering
%     \includegraphics[width=5.5cm]{imgs/accuracy_each_region.pdf}
%     %\vspace{-0.8cm}
%     \caption{Accuracy of \\ each region (our dataset)}
%     \label{fig:accuracy_each_region}
%     %\vspace{-0.3cm}
%   \end{minipage}
%   \begin{minipage}[t]{0.31\linewidth}
%     \centering
%     \includegraphics[width=5.5cm]{imgs/one_class_miss_each_region.pdf}
%     %\vspace{-0.8cm}
%     \caption{Accuracy of \\ each region (open dataset)}
%     \label{fig:two_class_miss_each_region}
%      %\vspace{-0.3cm}
%   \end{minipage}
% \end{figure*}

\subsubsection{Performance Stability}
\label{sec:performance-stability}
In this section, we assess the robustness of \sys \ to preserve the performance of region-specific taxi demand prediction in region-agnostic settings.
It is well-known that making the model more general by eliminating specialized components (i.e., region-dependent) from input leads to a significant drop in accuracy.
Therefore, we compare the performance of the model trained and tested using the same region 80-20 strategy, referred to as the "region-specific" approach\footnote{In this model, the region classifier is deactivated as it is trained with only one region.}, with the model trained using a leave one-city-out strategy, known as the "region-agnostic" approach.
Fig.~\ref{fig:accuracy_baseline} illustrates the accuracy comparison between the two test cases. The results indicate that the region-agnostic model achieves nearly identical prediction accuracy compared to the region-specific model, with a difference of less than $1\%$. This outcome can be attributed to the incorporation of a task-specific loss and excitation loss during the training of the \sys . These loss functions penalize the model for disentangling region-dependent components while simultaneously maintaining the accuracy of demand prediction.

% The results shown in Fig. \ref{fig:double_dataset_leave_one_out}, \ref{fig:accuracy_baseline} show that our proposed method achieves higher accuracy in the unseen region and is comparable to the region-specific model in terms of accuracy.

% \begin{figure}[tb]
%     \centering
%     \includegraphics[width=8cm]{imgs/accuracy_baseline.pdf}
%     \caption{Demand prediction accuracy}
%     \label{fig:accuracy_baseline}
% \end{figure}

% \begin{figure}[tb]
%     \centering
%     \includegraphics[width=8cm]{imgs/two_class_baseline_opendata.pdf}
%     \caption{Caption}
%     \label{fig:two_class_baseline_opendata}
% \end{figure}

\subsubsection{Performance Generalization}
\label{sec:performance-gener}
This section evaluates the system's ability to consistently predict taxi demand across diverse regions with varying geographical structures and mobility patterns. Fig.~\ref{fig:accuracy_each_region} and Fig.~\ref{fig:two_class_miss_each_region} depict the accuracy of \sys \  in different cities and datasets.
From Fig.~\ref{fig:accuracy_each_region}, our approach demonstrates consistent accuracy in predicting taxi demand across four regions ($r_1$, $r_2$, $r_3$, and $r_5$). Notably, region $r_4$ exhibits slightly higher accuracy due to its distinct characteristics, such as a homogeneous distribution of taxi demand patterns and minimal external factors influencing demand. These factors contribute to improved predictive accuracy compared to other regions. Similarly, Fig.~\ref{fig:two_class_miss_each_region} shows comparable accuracy in NYC, Chicago, and San Francisco.
These results indicate the generalizability of \sys , which effectively learns transferable feature representations to predict taxi demand across diverse regions.

\color{black}
To further clarify the decision logic of \sys\ and to verify whether the learned representations are properly disentangled into region-specific and region-agnostic components, we conduct an additional analysis focusing on how the model separates and utilizes regional characteristics. Specifically, the latent representation is decomposed into a region-specific embedding $z_r$ and a region-agnostic embedding $z$.
We evaluate this decomposition from two complementary perspectives: (i) the statistical independence between the two representations and (ii) the amount of regional information retained in each component.

To assess the independence between $z_r$ and $z$, we employ correlation-based metrics (corr) and Centered Kernel Alignment (CKA)~\cite{kornblith2019similarity}.
The correlation metric captures dimension-wise linear dependencies between the two embeddings, while CKA evaluates the similarity of their global representational structures.
The evaluation results for each dataset are summarized in Table~\ref{tab:disentanglement_eval}. As shown in the table, the CKA-based independence scores ($1-\mathrm{CKA}$) are high (0.99 for the open dataset and 0.95 for our dataset), indicating that $z_r$ and $z$ exhibit substantially different representational structures.
In addition, the correlation values are consistently low (0.05 and 0.12), suggesting minimal dimension-wise dependency between the embeddings. Together, these results confirm that the two representations capture largely distinct and complementary information, supporting effective disentanglement of region-specific and region-agnostic characteristics.

We further quantify how much regional information is encoded in each representation through a region-probing analysis. Specifically, we train a linear probe to predict region identifiers using each embedding independently. The results show that region classification can be performed with high accuracy using $z_r$, whereas the performance drops markedly when using $z$.
On average, the region-probe accuracy decreases, yielding an accuracy gap of approximately 0.30.
This clear gap indicates that regional information is predominantly captured by the region-specific embedding, while being largely suppressed in the region-agnostic one.
Importantly, despite suppressing region-specific signals, the region-agnostic embedding $z$ maintains strong predictive performance on the downstream task. This effect is particularly evident under leave-region-out evaluation settings, demonstrating that \sys\ preserves region-agnostic information that generalizes well to unseen regions. These findings indicate that \sys\ does not simply ignore inter-regional heterogeneity; rather, it explicitly isolates region-dependent factors while retaining transferable region-agnostic features. Through this structured disentanglement, \sys\ is able to adapt to significant regional differences while achieving robust generalization to previously unseen regions.

\color{black}

\begin{table}[t]
\centering
\caption{\textcolor{black}{Evaluation of Representation Disentanglement across Datasets.}}
\label{tab:disentanglement_eval}
\renewcommand{\arraystretch}{1.25}
\setlength{\tabcolsep}{10pt}
\resizebox{\linewidth}{!}{%
\begin{tabular}{lcc}
\toprule
\textbf{Metric} & \textbf{Open Dataset} & \textbf{Our Dataset} \\
\midrule
Correlation (corr) $\downarrow$ 
& 0.05 
& 0.12 \\

Independence (1--CKA) $\uparrow$ 
& 0.99 
& 0.95 \\

Region-Probe Gap $\uparrow$ 
& \makecell[c]{0.28 (0.81 $\rightarrow$ 0.52)} 
& \makecell[c]{0.30 (0.79 $\rightarrow$ 0.49)} \\

\bottomrule
\end{tabular}
}
\end{table}

\color{black}

\subsection{Comparative Evaluation}\label{sec:comparison_baseline}
In this section, we compare \sys \  with the most relevant state-of-the-art techniques:
 Graph neural network (GCN),  
The Deep Multi-View Spatial-Temporal Network (DMVST-Net)\cite{yao2018deep}, Node2Vec\cite{10.1145/2939672.2939754}, Spatial-Temporal Graph Auto-Encoder (STGAE) \cite{9583812}, the Multi-view Spatial Network Embedding (MSNE) \cite{fu2019efficient}, Ada-STGCN~\cite{yao2023transfer}, ST-DAN\cite{wang2021spatio}.

GCN\cite{kipf2017semisupervised} is a traditional method, and we fed a semantic graph into GCN to represent the input data.
DMVST-Net \cite{yao2018deep} is a deep learning framework proposed for taxi demand prediction, which leverages multi-view information. DMVST-Net comprises three distinct views: the temporal view, the spatial view, and the semantic view, which are modeled using LSTM, CNN, and graph embedding, respectively.
Node2Vec\cite{10.1145/2939672.2939754} is an unsupervised node embedding method that learns continuous feature representations for nodes. We employ Node2Vec to obtain node representations from each region graph and then train an MLP model to predict taxi demand based on the obtained node representations.
STGAE\cite{9583812} employs  Graph Auto-Encoder to extract low-dimensional latent representations from spatio-temporal graphs by minimizing the graph reconstruction loss.
MSNE\cite{fu2019efficient} is a multi-view based node embedding method that incorporates an inter-region autocorrelation layer to capture inter-region similarities. 
\textcolor{black}{
Ada-STGCN~\cite{yao2023transfer} is a graph-based transfer learning model that applies adversarial domain adaptation to learn domain-invariant node embeddings across source and target cities, which are subsequently used to model spatio-temporal dynamics.
ST-DAN\cite{wang2021spatio} is Spatial-temporal knowledge transfer via deep adaptation networks, which is a domain
adaptation method that adopts the STGCN model to
learn the source and target representations and reduces the domain discrepancy by minimizing the MMD (Maximum Mean Discrepancy) loss.
MMD is a statistical measure of the distance between two data distributions; reducing MMD encourages the model to produce similar feature distributions across cities, thereby preventing the learned representations from overfitting to region-specific patterns.
}

\begin{figure*}[!t]
    \centering
    \includegraphics[width=15cm]{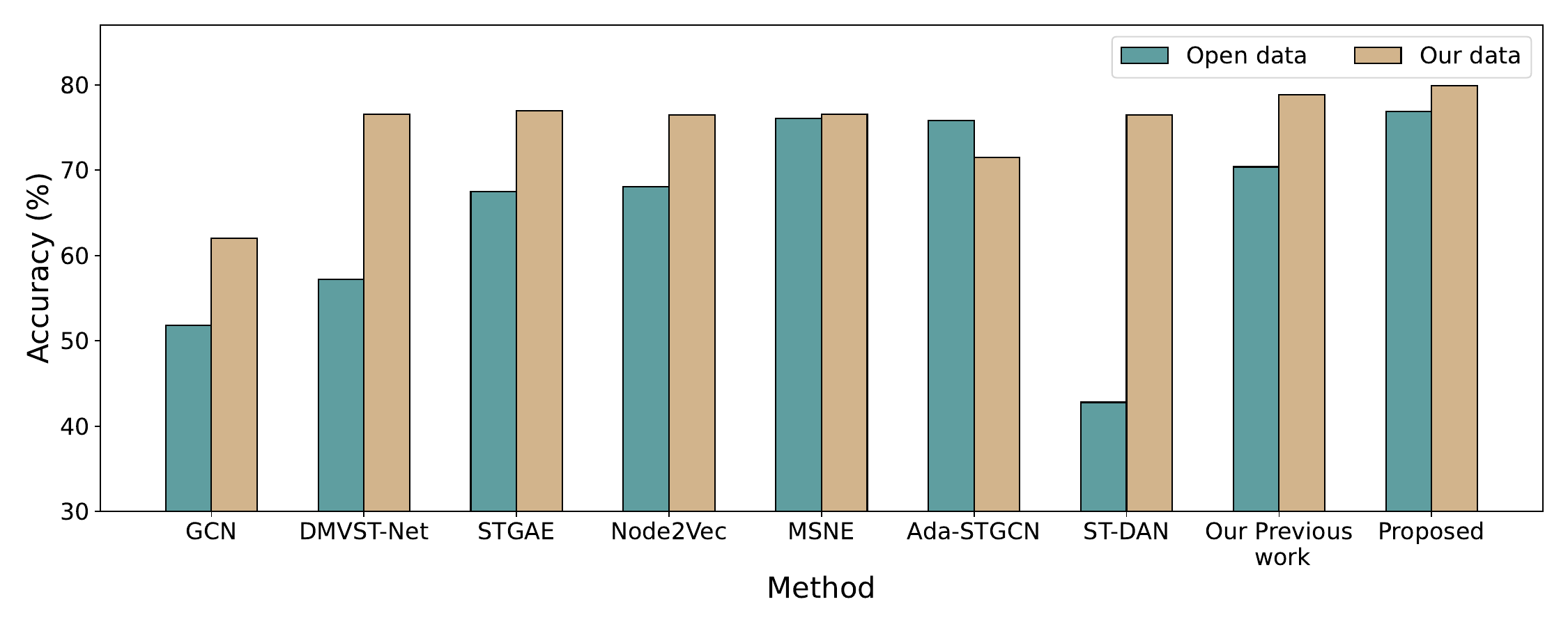}
    \caption{Demand prediction accuracy of each method.}
    \label{fig:double_dataset_leave_one_out}
\end{figure*}
Fig.~\ref{fig:double_dataset_leave_one_out} illustrates the performance of \sys \  in comparison to the existing approaches. 
The results demonstrate the effectiveness of \sys \  and its ability to achieve an accuracy of 80.2\%.
With this accuracy, it outperforms the existing methods by up to \textcolor{black}{28.6}\%. Notably, our approach exhibits superiority even when compared to the current state-of-the-art techniques, i.e., MSNE, Node2Vec, and Ada-STGCN by a minimum margin of 3.5\%.
\textcolor{black}{Furthermore, our method consistently achieves higher prediction accuracy than ST-DAN, which is an MMD-based domain alignment approach, as observed in Fig.~\ref{fig:double_dataset_leave_one_out}.
While ST-DAN encourages distribution alignment across regions, applying it to a single spatio-temporal view may lead to excessive alignment and limit the preservation of task-relevant regional characteristics.
In contrast, \sys\ performs feature extraction by leveraging multi-view graph information and employs a framework based on VAE and adversarial learning to disentangle region-agnostic representations while preserving as much task-relevant information as possible.
As a result, the model maintains high generalization performance even on unseen regions.}
The success of our approach can be attributed to its unique capability to capture \textcolor{black}{region-agnostic} representations.
This feature allows our taxi demand prediction approach to generalize effectively to unseen regions.
In contrast, other methods rely solely on neural networks and multi-view representational abilities, or they emphasize similarities between different regions. However, these approaches cannot ensure general representation and dependable cross-region performance.

\begin{figure*}[t]
  \centering

  \begin{subfigure}[t]{0.24\textwidth}
    \centering
    \includegraphics[width=\linewidth]{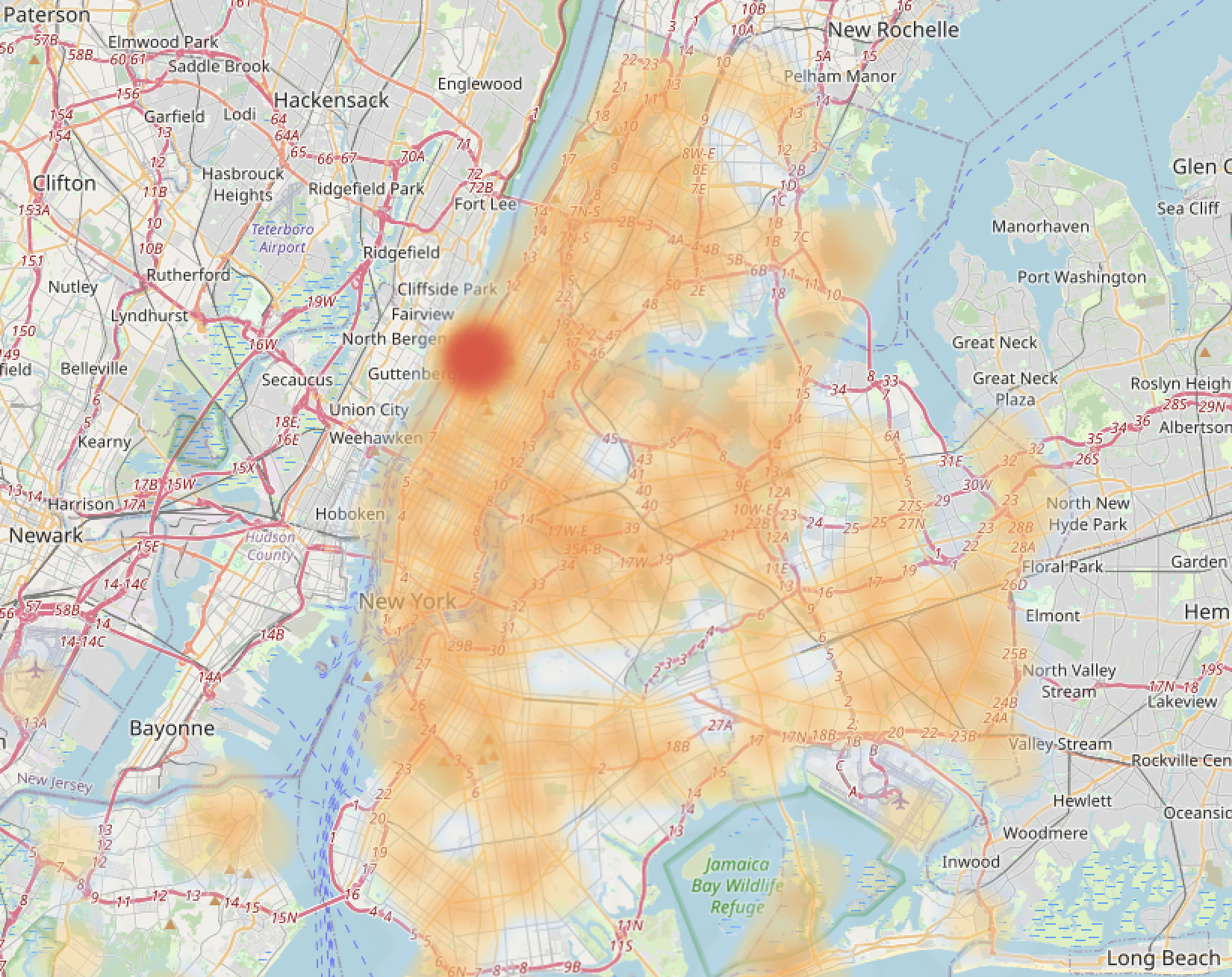}
    \caption{ST-DAN\cite{wang2021spatio}}
  \end{subfigure}\hfill
  \begin{subfigure}[t]{0.24\textwidth}
    \centering
    \includegraphics[width=\linewidth]{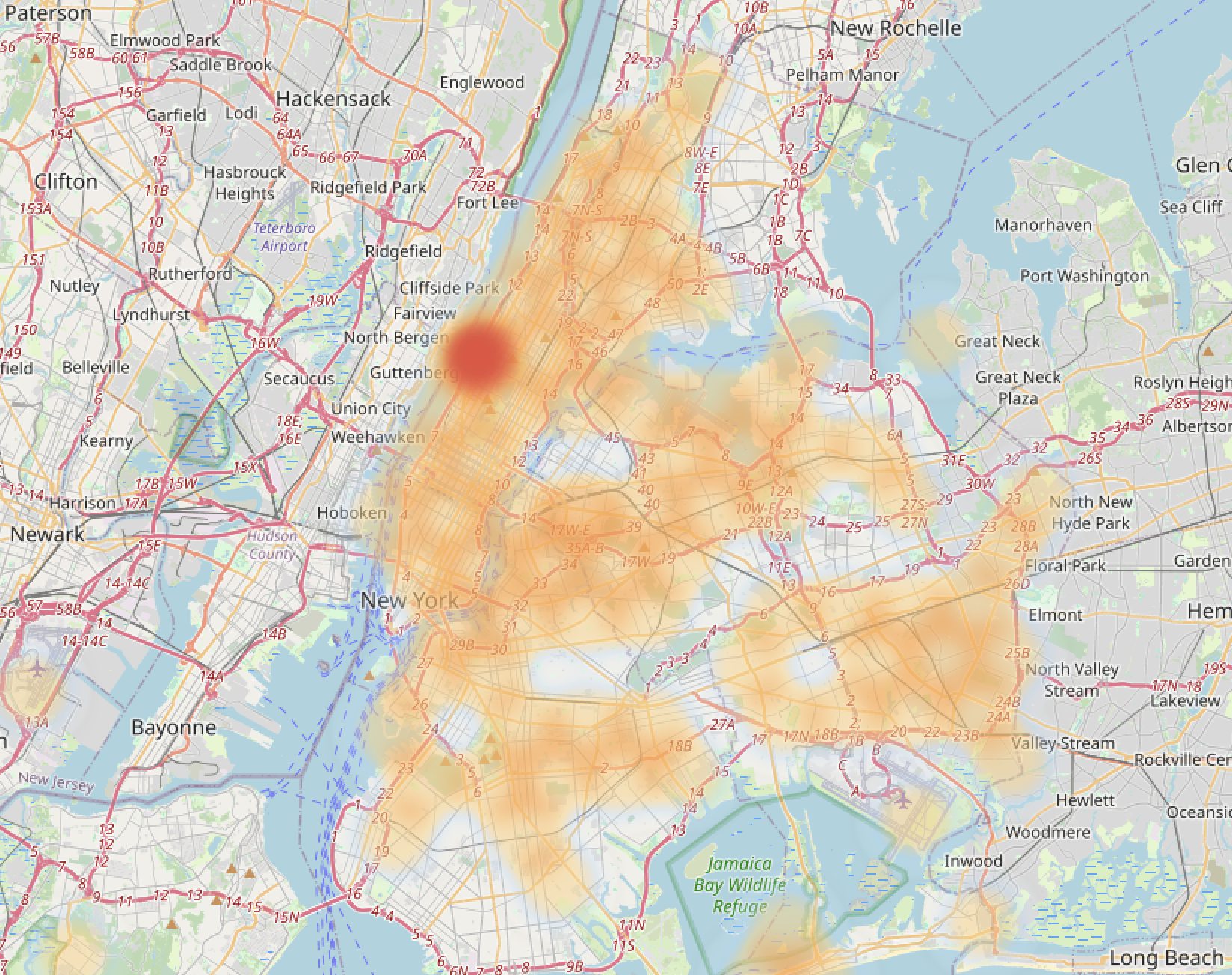}
    \caption{Our Previous Method}
  \end{subfigure}\hfill
  \begin{subfigure}[t]{0.24\textwidth}
    \centering
    \includegraphics[width=\linewidth]{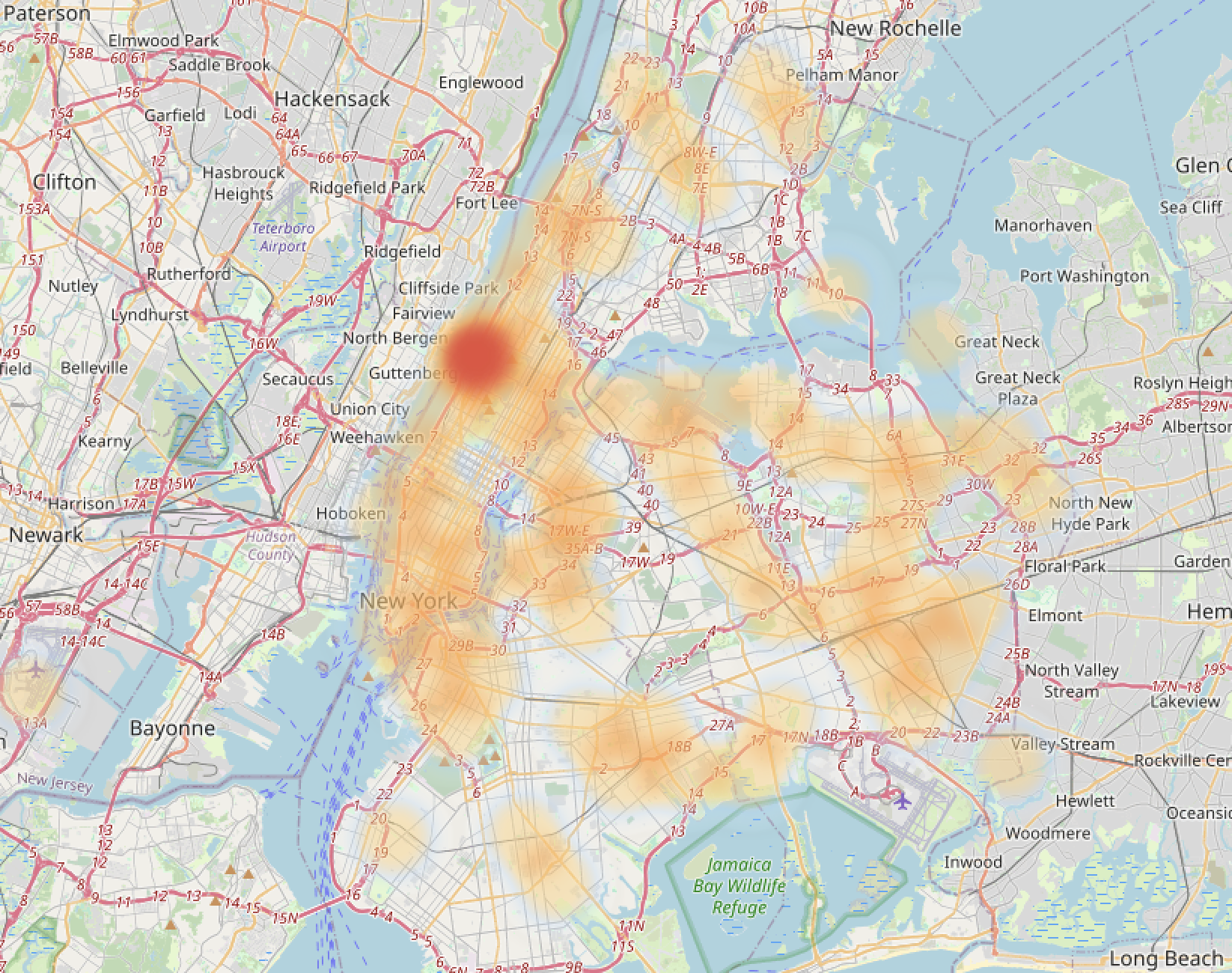}
    \caption{Ada-STGCN\cite{yao2023transfer}}
  \end{subfigure}\hfill
  \begin{subfigure}[t]{0.24\textwidth}
    \centering
    \includegraphics[width=\linewidth]{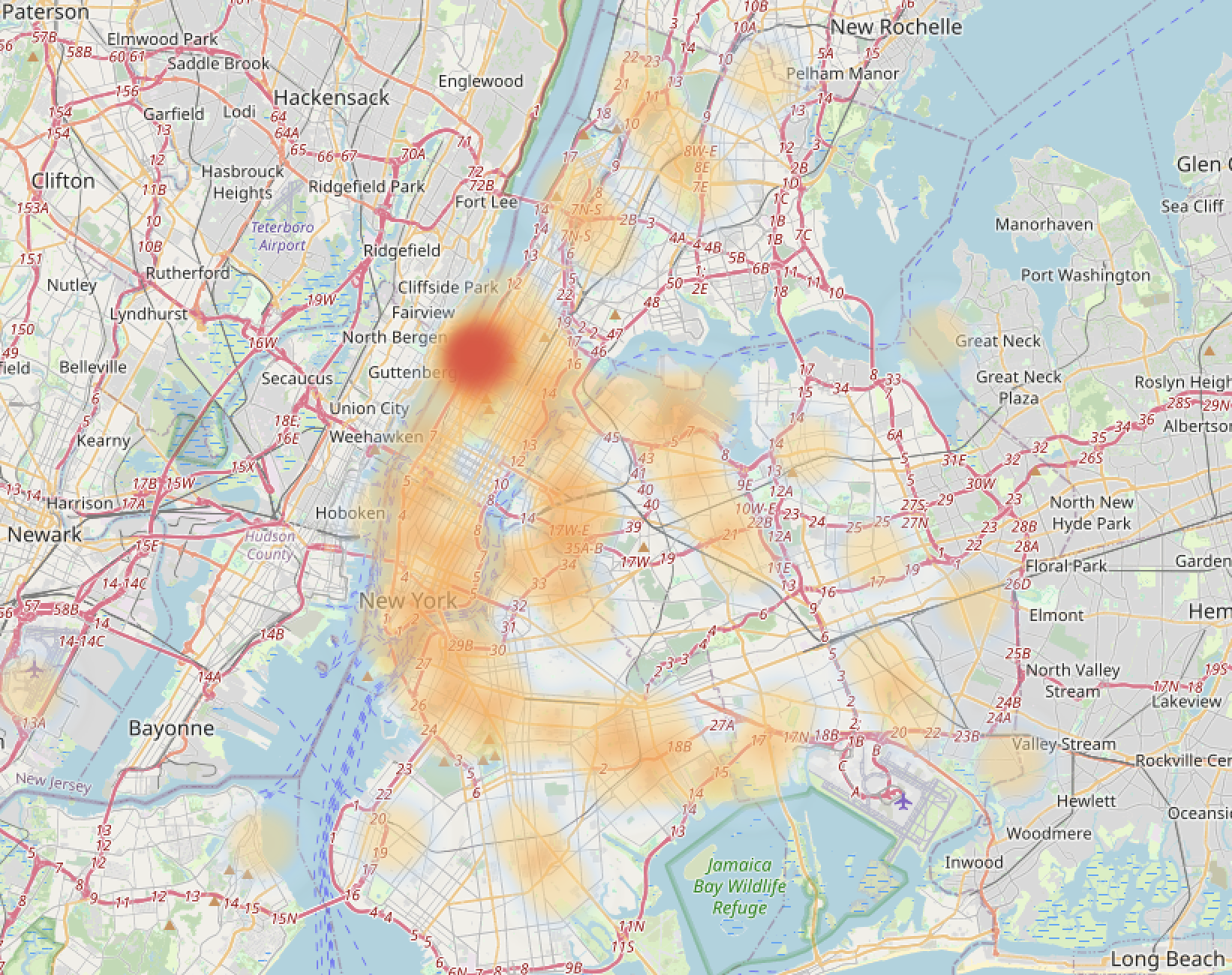}
    \caption{Proposed}
  \end{subfigure}

  \caption{\textcolor{black}{Spatial visualization of prediction errors on unseen regions.
The maps show the absolute difference between ground-truth demand and predicted demand for the proposed method and baseline approaches. Darker colors indicate larger prediction errors.}}
  \label{fig:four_in_a_row}
\end{figure*}

\textcolor{black}{
In addition to quantitative metrics, we further evaluate the proposed method through visual inspection to assess its prediction behavior on unseen regions.
Specifically, we visualize the spatial distribution of prediction errors produced by the proposed method and other relevant models on regions that were not observed during training.
Fig. \ref{fig:four_in_a_row} illustrates the absolute difference between the ground-truth demand and the predicted demand for each spatial cell, where darker colors indicate larger prediction errors.
This visualization enables intuitive comparison of how prediction errors are distributed over space.
As shown in fig.\ref{fig:four_in_a_row}, the proposed method yields consistently smaller and less dispersed errors compared to the baseline approaches.
These results further support the effectiveness of \sys\ in achieving robust generalization across regions.
}

\subsection{Performance Stability on Domain Shift}

\begin{table}[t]
\centering
\caption{Evaluation of the proposed system under two domain shift settings: \emph{region shift} and \emph{temporal shift}.}
\label{tab:domain_shift_eval}
\renewcommand{\arraystretch}{1.25}
\setlength{\tabcolsep}{9pt}
\begin{tabular}{lcc}
\toprule
\textbf{Domain Shift} & \textbf{Open Dataset} & \textbf{Our Dataset} \\
\midrule

\makecell[l]{Region Shift \\ (train regions $\rightarrow$ unseen regions)}
& 0.798 
& 0.769 \\

\midrule

\makecell[l]{Temporal Shift \\(past $\rightarrow$ future)}
& 0.768 
& 0.536 \\

\midrule
\makecell[l]{Random split \\ (no region and temporal shift)}
& 0.855 
& 0.799 \\

\bottomrule
\end{tabular}
\end{table}

\color{black}
The proposed method is primarily designed for cross-region transferability, but its architecture also suggests potential robustness to other forms of domain shift.
While Section \ref{sec:comparison_baseline} evaluates generalization under regional distribution shifts, we additionally examine temporal domain shifts to assess broader model stability.
We consider three evaluation settings: (i) region shift, where the model is trained on a subset of regions and tested on unseen regions; (ii) temporal shift, where the model is trained on past data and evaluated on future periods; and (iii) a random split as a baseline.

As shown in Table \ref{tab:domain_shift_eval}, on the open dataset, the proposed method achieves comparable performance under regional and temporal shifts as well as the random split, indicating robustness to multiple types of distributional change.
In contrast, in our collected dataset, performance degrades more under temporal shift than under regional shift.
Compared to the open dataset, our dataset is derived from smaller-scale operators with more limited regional coverage, resulting in lower demand density per unit time. 
While the open dataset contains approximately 300K trips within a month, our dataset includes only about 30K trips despite spanning more than five months. 
Under the temporal shift setting, this sparse and less comprehensive coverage leads to a larger domain shift, which causes performance degradation on our dataset.
% We expect that this limitation can be alleviated by training on a larger dataset.

\begin{comment}
Compared to the open dataset, our dataset is derived from smaller-scale operators with more limited regional coverage, resulting in lower demand density per unit time. 
While the open dataset contains approximately 300K trips within a month, our dataset includes only about 30K trips despite spanning more than five months. 
Under the temporal shift setting, this sparse and less comprehensive coverage leads to a larger domain shift, which causes performance degradation on our dataset.
\end{comment}

\color{black}

%\vspace{-0.1cm}
\section{Related Work} \label{sec:related_work}
A concise comparison of our approach and related studies is presented in Table \ref{tab:related_work_summary}.
\begin{table*}[tb]
    \centering
    \caption{Related work summary.}
    \begin{tabular}{l|ccccc}
    \hline
        ~ & 
        \begin{tabular}{c}
            Taxi demand \\ prediction
        \end{tabular}
        &
        \begin{tabular}{c}
            Multi-view\\Spatio-temporal\\prediction
        \end{tabular}
        &
        \begin{tabular}{c}
            Transferable to \\ other regions
        \end{tabular}
        &
        \begin{tabular}{c}
            Hexagonal gridding
        \end{tabular}
        &
        \begin{tabular}{c}
            Region-Agnostic \\ feature extraction
        \end{tabular} \\
        \hline
        \hline
        \textbf{Our work} &  \checkmark & \checkmark & \checkmark & \checkmark  & \checkmark \\
        \textit{Our previous work} \cite{ozeki2023onemodel} & \checkmark & & \checkmark & \checkmark & \checkmark \\
        Zhang, Daqing, et al. \cite{Understanding_Taxi_Service_Strategies} & \checkmark & & & &\\
        STAR \cite{safikhani2020spatio} & \checkmark & & & &\\
        Xiang Yan, et al. \cite{yan2020using} & \checkmark & & & &\\
        Chen, Zhe, et al. \cite{chen2020multitask} & \checkmark & \checkmark & & &\\
        GCN \cite{kipf2017semisupervised} & & & &\\
        RNN taxi demand prediction \cite{xu2017real} & \checkmark & & & &\\
        ARIMA-base model \cite{8257080} & \checkmark & & & &\\
        MLRNN \cite{9439926} & \checkmark & & & &\\
        DMVST-Net \cite{yao2018deep} & \checkmark & \checkmark & & & \\
        STL \cite{wang2018stratified} &  & \checkmark & \checkmark & & \\
        Node2Vec \cite{10.1145/2939672.2939754} &  &  & \checkmark & \\
        STGAE \cite{9583812} & & & \checkmark &  & \\
        MSNE \cite{fu2019efficient} & & \checkmark & \checkmark & & \\
        CGAL \cite{zhang2019unifying} & & \checkmark & \checkmark & & \\
        MIFGNN \cite{liao2022taxi} & \checkmark & \checkmark & &  & \\
        \color{black} MVSTD \cite{FAN2025114161} & \checkmark & \checkmark & & &\\
        H-CNN \cite{ke2018hexagon} & \checkmark & &  & \checkmark  & \\
        OGCRNN \cite{guo2020optimized} & \checkmark & &  & & \\
        MVGCN \cite{sun2020predicting} &  & \checkmark & & & \\
        Yu, Byeonghyeop et al. \cite{yu2020forecasting} & & & & & \\
        ToGCN \cite{qiu2020topological} & & & & & \\
        T-GCN \cite{Zhao_2020} & & & & & \\
        GCN-DHSTNet \cite{ali2022exploiting} & & & & & \\
        TFCS \cite{coston2019fair} & & & \checkmark & & \\
        Kimura, Nobuaki, et al. \cite{kimura2019convolutional} & & \checkmark & \checkmark & & \\
        \color{black} HGMN \cite{JI2026130316} & & \checkmark & & &\\
        \color{black} SFMGTL \cite{CHEN2024104604} & \checkmark & \checkmark & \checkmark & &\\
        \color{black} MDTLGCN \cite{SHAO2025105152} & & \checkmark & \checkmark & & \\ 
        \color{black} CRRL \cite{DAI2025103215}  & & \checkmark & \checkmark & & \checkmark \\ \hline
    \end{tabular}
    \label{tab:related_work_summary}
\end{table*}
\color{black}

\subsection{Taxi Demand Prediction} 
\color{black}
The prediction of taxi demand has recently garnered considerable attention, owing to the abundance of large-scale spatio-temporal data that facilitates the training of deep neural networks, such as Convolutional Neural Networks (CNNs) and Long Short-Term Memory (LSTM) networks.

Recent studies have leveraged neural networks to predict taxi demand with greater accuracy\cite{ourMDMpaper,  gotoprivacy,acm_src}.
For example, the method proposed in \cite{yao2018deep} employs a CNN to capture spatial features and an LSTM to capture temporal features, resulting in improved accuracy compared to methods that only consider semantic, spatial, or temporal information.
The author of \cite{9172100} improves accuracy by leveraging spatio-temporal correlations between pick-up and drop-off locations through multitask learning.
Other studies have focused on accounting for the heterogeneity of taxi demand across zones within a region\cite{9439926}.
MLRNN (MultiLevel Recurrent Neural Networks) proposes in \cite{9439926} clusters target region into multiple clusters according to their correlations to taxi demand data and trains zone-specific models to predict demand, taking into account the unique distribution and temporal variability of demand in each zone. 
% While these machine learning-based methods have shown promising results when applied to spatio-temporal data, they do not consider privacy implications.
% \textcolor{blue}{While these machine learning-based methods have shown promising results when applied to spatio-temporal data, they typically assume unrestricted access to fine-grained mobility traces and thus pay limited attention to privacy implications. From a transportation data perspective, the privacy vulnerability of individual mobility traces and the associated privacy–utility trade-off have been quantitatively investigated, highlighting that releasing trajectory-like records can entail non-trivial re-identification risks\cite{gao2019quantifying}.}
The author of \cite{liao2022taxi} integrated semantic data that represent external factors influencing road conditions like weather, urban events, and their text descriptions to improve the accuracy of taxi demand prediction.
The authors apply the multimodal attention mechanism (MIFGNN) to extract spatial and temporal features and boost the prediction model.
As a result, experiments demonstrate an improvement of MIFGNN over all baselines is statistically significant with P-Value $< 0.05$ under the Mann-Whitney U test and state that the usage of multi-source information improves taxi demand forecasting. 
\textcolor{black}{MVSTD\cite{FAN2025114161} introduces a multi-view spatio-temporal graph framework that explicitly extracts influence features from external factors, such as weather, air quality, and public holidays, using 3D-CNNs to denoise historical data.}
% These machine learning-based methods have shown promising results.
\textcolor{black}{In parallel, recent studies have also examined downstream operational decisions in ride-hailing systems that rely on spatio-temporal demand–supply signals. For instance, deep reinforcement learning has been used to proactively dispatch vacant vehicles toward regions with large demand gaps\cite{LIU2022102694}, and deep multi-task learning has been proposed to dynamically determine matching radii for on-demand ride services under multiple system objectives\cite{CHEN2025103822}.}

\textit{However, it is difficult to transfer these models to another unseen region because these methods focus on accurately predicting taxi demand within the target region.
Different from these existing methods, \sys\  aims to predict taxi demand accurately even in unseen regions.}

% As a result, experiments demonstrate an improvement of MIFGNN over all baselines is statistically significant with P-Value < 0.05 under the Mann-Whitney U test and state that usage of multi-source information improves taxi demand forecasting. 
% These machine learning-based methods have shown promising results, however, they simply extracted spatio-temporal data with LSTM and they did not fully explore dynamic dependencies of spatio-temporal view, semantic view, and mobility view together. 

\subsection{Spatial-Temporal data processing using Graph Neural Network}
In traffic prediction problems, there are many kinds of applications to predict any traffic-related data, such as traffic volume (collected from GPS or loop sensors), traffic flow, and taxi demand (our problem). 
The problem formulation process for these different types of traffic data is the same in terms of spatio-temporal application.
Essentially, the goal is to predict a traffic-related value for a location at a timestamp.

% In order to predict traffic-related value, time series prediction methods have been widely used.
% For example, Autoregressive Integrated
% Moving Average models (ARIMA) and LSTM are basic methods to predict time series value and are used for traffic prediction\cite{alghamdi2019forecasting, yang2019traffic, ran2019lstm}.
% Although this time series forecasting method can capture temporal correlation, 
To capture spatial and temporal correlation simultaneously and improve the accuracy of traffic-related prediction, some research employed
CNN (convolutional neural network) and GCN (Graph convolutional network) \cite{ke2018hexagon, guo2020optimized, sun2020predicting}.
Since CNNs were designed for Euclidean space, such as images and grids, they have limitations in transportation networks with non-Euclidean topology and thus cannot essentially characterize the spatial correlation of traffic flow and road network.
On the other hand, graph convolutional neural networks (GCNs) are dedicated to processing network structures\cite{yu2020forecasting, qiu2020topological}, which can better model the spatial dependence of road segments on traffic networks\cite{Zhao_2020}.
\textcolor{black}{For example, HGMN \cite{JI2026130316} utilizes a dynamic memory graph network within a lightweight U-Net-based framework to capture both short-term fluctuations and long-term periodic patterns through multi-scale feature extraction.}

Since traffic-related value also depends on human activity and other external factors, prediction based on historical spatial information alone has limits because these methods cannot consider them, such as Point of Interest(POI) and weather.
In fact, some previous works improve the accuracy of traffic-related prediction by leveraging multi-view information that includes POI and weather \cite{10.1145/2424321.2424340, ali2022exploiting}.
This multi-view information is expected to be important when the model is used in the unseen region because spatio-temporal traffic pattern depends on each region.

\textit{Based on this analysis of existing work, we employ multi-view graph processing modules in the proposed framework.
Through the experiments in Section \ref{sec:multi_view_impact}, we confirm that this multi-view graph processing contributes to the performance of the model in the unseen region.}

\subsection{Spatial-Temporal data processing on cross-region}
% Recently, many papers have been introduced for intelligent transport systems to solve prediction problems such as time travel, crowd flow, taxi demand, and trajectory forecasting \cite{xu2023dynamic, poongodi2022new, KE2021102858, li2023dynamic}. Although many of these methods demonstrated decent and accurate results, prediction models are trained on large-scale data. 

The technique to realize a cross-region machine learning model can be roughly grouped into two categories: transfer learning and representation learning.
Machine learning models for spatio-temporal applications do not work well in regions where the data is scarce.
All data-driven models face the challenge of obtaining accurate results without increasing excessive input and preserving the heterogeneity of data. 
Much research addresses this problem by transfer learning which is a method to transfer knowledge from regions with lots of data to regions where the data is scarce\cite{liu2023real,wang2018stratified, coston2019fair,kimura2019convolutional}. 
% \cite{liu2023real,wang2018stratified, coston2019fair,kimura2019convolutional}
In spatio-temporal applications, transfer learning successfully produces results.
\textcolor{black}{Recent advancements like SFMGTL\cite{CHEN2024104604} utilize hierarchical node clustering to capture multi-granularity patterns, while MDTLGCN \cite{SHAO2025105152} employs knowledge distillation to compensate for severe missing data by extracting city-agnostic features. Furthermore, CRRL \cite{DAI2025103215}  introduces contrastive learning to identify reliable region-level correspondences between cities, mitigating negative transfer. However, transfer learning still requires some labeled data or historical records in target regions for alignment, which limits its practicality in entirely unobserved areas.}
% However, transfer learning needs labeled data in target regions.
% Therefore, transfer learning is not necessarily practical and applicable to regions where there is no labeled data.

Another technique to realize a cross-region machine learning model is graph representation learning.
Graph representation learning aims to find a good mapping function to represent the location within a region into low-dimensional latent embedding. 
Since a good data representation is useful for multiple downstream tasks, the model trained by such embedding is expected to be less affected by region-specific knowledge.
% can alleviate data collecting by reducing the volume of processed data.   
One of the network embedding methods Node2Vec \cite{10.1145/2939672.2939754} aims to learn the latent representation of a node to capture local structures by random walks.
However, this method uses a shallow network and cannot contain node feature information.
To consider node feature information, a GCN-based representation learning method is proposed \cite{fu2019efficient, 9583812}. 
Spatio-temporal graph autoencoder (STGAE) extracts spatio-temporal graph representation that is useful for multi-task \cite{9583812}.
\textcolor{black}{We include graph autoencoder-based baselines in our experiments. Related temporal graph autoencoding ideas have also been explored for travel and origin–destination forecasting, such as TGAE \cite{wang2022tgae}.}
Multi-view Spatial Network Embedding (MSNE) \cite{zhang2019unifying} is unsupervised spatial representation learning via dual-adversarial auto-encoder framework using multi-view graphs.
These existing graph representation learning successfully extract effective representation for spatial application within the region where the feature extractor was trained.
However, these existing representation learning methods do not necessarily lead to a cross-region model performance because these methods only consider graph reconstruction loss, and cannot guarantee region independence of latent features and the utility of the target task.

\textit{In contrast, our proposed system focuses on the idea of one model for all regions that process multi-view graphs with a novel region agnostic mechanism consisting of two encoders responsible for extracting  \textcolor{black}{region-agnostic} and region-specific features.
This method allows for predicting taxi demand in unseen regions with high accuracy.
}

\section{Conclusion} \label{sec:conclusion}
In this paper, we have introduced a taxi demand prediction framework that effectively forecasts taxi demand even in previously unseen regions by leveraging a region-agnostic mechanism. The proposed framework \sys \  incorporates a multi-view GCN to represent the input spatiotemporal, mobility, and semantic features of the region. Additionally, we have proposed a \textcolor{black}{region-agnostic} feature extraction mechanism utilizing two independent encoders to separate \textcolor{black}{region-agnostic} and region-specific factors. 
We conducted experiments using well-known real-world open datasets and our own dataset. The results demonstrate that our approach achieves the highest accuracy in taxi demand prediction for unseen regions compared to state-of-the-art approaches. We also conducted a module-level analysis, evaluating the impact of hexagonal virtual gridding, multi-view graph processing, and the region-agnostic module. The results indicate that each module contributes to the extraction of \textcolor{black}{region-agnostic} features and leads to high accuracy in taxi demand prediction for unseen regions.
These findings demonstrate the feasibility and practicality of achieving accurate predictions in real-world applications, thereby paving the way for enhanced decision-making, resource allocation, and operational efficiency in the taxi service industry.
\textcolor{black}{
A limitation of CROSS-Net is that disentanglement between region-specific and region-agnostic latent features is not enforced by explicit constraints such as orthogonality or mutual-information regularization.
Exploring more rigorous disentanglement mechanisms and other stochastic optimization strategies, such as SGLD\cite{welling2011bayesian}, is an important direction for future work and may further improve model robustness and interpretability.
}

\color{black}
% \section*{Acknowledgment} 
% % %\vspace{-0.2cm}
% This work was partially funded by JST, CREST Grant JPMJCR21M5, and JSPS, KAKENHI Grant 22K12011, and NVIDIA award.

\bibliographystyle{IEEEtran}
\bibliography{IEEE_ref}

\end{document}